\documentclass[12pt]{article}%
\usepackage{amsmath}
\usepackage{amsfonts}
\usepackage{amssymb}
\usepackage{longtable}
\usepackage{multirow}
\usepackage{rotating}
\usepackage{graphicx}%
\usepackage{color}
\usepackage{graphics}
\usepackage{graphicx}
\usepackage{epsfig}
\usepackage{subfigure}
\usepackage{latexsym,bm}
\usepackage{graphicx}%
\usepackage{mathrsfs}
\usepackage{cite}
\usepackage{mathtools}
\usepackage{url}
\usepackage{hyperref}
\usepackage{diagbox}
\usepackage{pifont}

\usepackage{algorithm}

\usepackage[usenames,dvipsnames]{xcolor}
\usepackage{tabularx}
\usepackage{array}
\usepackage{colortbl}
\usepackage[T1]{fontenc}
\usepackage[utf8]{inputenc}

\newtheorem{example}{Example}

\begin{document}

\def\R{{\mathbb R}}
\newcommand{\bbb}[1]{{\boldsymbol  #1 }}
\newcommand{\vth}{{\vartheta}}
\newcommand{\blue}[1]{{\color{blue} #1 }}
\newcommand{\red}[1]{{\color{red} #1 }}
\newcommand{\magenta}[1]{{\color{magenta} #1 }}
\newcommand{\g}[1]{{\color{teal} #1 }}

\newcommand{\eff}{$\kappa_{\rm eff}$}
\title{Power-Enhanced Residual Network for Function Approximation and Physics-Informed Inverse Problems}

%\author{
%Amir Noorizadegan\footnotemark[1] {Department of Civil Engineering, National Taiwan University, 10617, Taipei, Taiwan}
%D.L. Young{Core Tech System Co. Ltd, Moldex3D, Chubei, Taiwan},{Department of Civil Engineering, National Taiwan University, 10617, Taipei, Taiwan}
%Y.C. Hon{Department of Mathematics, City University of Hong Kong, SAR, Hong Kong, China}
%C.S. Chen{Department of Civil Engineering, National Taiwan University, 10617, Taipei, Taiwan}
%}

\author{
A. Noorizadegan\thanks{Department of Civil Engineering, National Taiwan University, 10617, Taipei, Taiwan}
, D.L. Young \footnotemark[1]
\thanks{Core Tech System Co. Ltd, Moldex3D, Chubei, Taiwan} \footnotemark[4]
, Y.C. Hon\thanks{Department of Mathematics, Chinese University of Hong Kong, SAR, Hong Kong, China}  \footnotemark[4], C.S. Chen\footnotemark[1]
 \thanks{Corresponding authors: \texttt{dchen@ntu.edu.tw}, \texttt{dlyoung@ntu.edu.tw}, \texttt{Benny.hon@math.cuhk.edu.hk}}
}

\date{}

\maketitle
%\today

\begin{abstract}

In this study, we investigate how the updating of weights during forward operation and the computation of gradients during backpropagation impact the optimization process, training procedure, and overall performance of the neural network, particularly the multi-layer perceptrons (MLPs). This paper introduces a novel neural network structure called the Power-Enhancing residual network, inspired by highway network and residual network, designed to improve the network's  capabilities for both smooth and non-smooth functions approximation in 2D and 3D settings. By incorporating power terms into residual elements, the architecture enhances the stability of weight updating, thereby facilitating better convergence and accuracy. The study explores network depth, width, and optimization methods, showing the architecture's adaptability and performance advantages.
Consistently, the results emphasize the exceptional accuracy of the proposed Power-Enhancing residual network, particularly for non-smooth functions. Real-world examples also confirm its superiority over plain neural network in terms of accuracy, convergence, and efficiency.
Moreover, the proposed architecture is also applied to solving the inverse Burgers' equation, demonstrating superior performance. In conclusion, the Power-Enhancing residual network offers a versatile solution that significantly enhances neural network capabilities by emphasizing the importance of stable weight updates for effective training in deep neural networks. The codes implemented are available at:
\url{https://github.com/CMMAi/ResNet_for_PINN}.

\end{abstract}

\noindent{Keywords: Residual network, Highway network for PINN, Neural networks, Partial differential equations, inverse problem, interpolation, function approximation.  }
\section{Introduction}

Deep neural networks have revolutionized the field of machine learning and artificial intelligence, achieving remarkable success in various applications, including image recognition, natural language processing, and  reinforcement learning. Moreover, their adaptability extends beyond these domains, as evidenced by their effective integration with physics-informed neural network (PINN) approaches \cite{Raissi19}.  Both theoretical insights and empirical observations consistently highlight the critical role of neural network depth in determining their effectiveness \cite{Srivastava15a}. As emphasized by Bengio and colleagues \cite{Bengio13}, employing deep networks can confer computational and statistical advantages for tackling complex tasks. However, achieving optimal performance with deeper networks is not a simple matter of adding layers. It has become evident that optimizing deep networks presents significant challenges, leading to investigations into various methodologies such as initialization strategies \cite{Glorot10,He15} or staged training approaches \cite{Romero14}.

A significant advancement in this field occurred with the introduction of the highway network \cite{Srivastava15a,Srivastava15} and its derivative, the residual networks, often referred to as ResNets \cite{He15,He16}. These innovations demonstrated exceptional performance in constructing deep architectures and addressing the issue of vanishing gradients. ResNets leverage skip connections to create shortcut paths between layers, resulting in a smoother loss function. This permits efficient gradient flow, thus enhancing training performance across various sizes of neural networks \cite{Li18}. Our research aligns closely with theirs, particularly in our exploration of skip connections' effects on loss functions.
In 2016, Veit et al. \cite{Veit16} unveiled a new perspective on ResNet, providing a comprehensive insight. Velt's research underscored the idea that residual networks could be envisioned as an assembly of paths with varying lengths. These networks effectively employed shorter paths for training, effectively resolving the vanishing gradient problem and facilitating the training of exceptionally deep models. Jastrz\k{e}bski et al.'s research \cite{JastrzSebski17} highlighted Residual Networks' iterative feature refinement process numerically. Their findings emphasized how residual connections guided features along negative gradients between blocks, and show that effective sharing of residual layers mitigates overfitting.

In related engineering work, Lu et al. \cite{LuLu20} leveraged recent neural network progress via a multifidelity (MFNN) strategy (MFNN: refers to a neural network architecture that combines outputs from multiple models with varying levels of fidelity or accuracy) for extracting material properties from instrumented indentation (see \cite{LuLu20}, Fig.~1(D) ).
The proposed MFNN in this study incorporates a residual link that connects the low-fidelity output to the high-fidelity output at each iteration, rather than between layers.
Wang et al. \cite{Wang21} proposed an improved fully-connected neural architecture. The key innovation involves integrating two transformer networks to project input variables into a high-dimensional feature space. This architecture combines multiplicative interactions between the plain network's outputs and residuals, resulting in improved predictive accuracy, but with a high cost of CPU time as reported.

% \red{This architecture uses a combination of plain network outputs with residual-based data, resulting in improved accuracy and stability. On the other hand, larger CPU time per epoch (iteration) are expected due to adding more computations to the network of proposed architecture. In spite of that, we observed that the proposed method may lead to a smaller CPU time for the entire training due to increase of faster convergence in comparison with the plain network.}

In this paper, we propose a novel architecture called the Power-Enhancing SkipResNet, aimed at advancing the interpolation capabilities of deep neural networks for smooth and non-smooth functions in 2D and 3D domains.
The key objectives of this research are as follows:
\begin{itemize}
    \item Introduces the ``Power-Enhancing SkipResNet'' architecture,
    \item Enhances network's expressive power for improved accuracy and convergence,
    \item Outperforms conventional plain neural networks,
    \item Conducts extensive experiments on diverse interpolation scenarios and inverse Burger's equation,
    \item Demonstrates benefits of deeper architectures and
    \item Investigate the effect of weight updating during forward operation and gradients during backpropagation on the optimization process, training procedure, and overall performance of the neural network. 
\end{itemize}
Through rigorous analysis and comparisons, we demonstrate the advantages of the proposed architecture in terms of accuracy and convergence speed.
The remainder of this paper is organized as follows: Section 2 reviews the neural network and its application for solving interpolation problems. In Section 3, we briefly presents {physics-informed} neural network for solving inverse Burgers' equation. Section 4 discusses the residual network and the proposed Power-Enhancing SkipResNet, explaining the incorporation of power terms and its potential benefits. Section 5 presents the experimental setup and the evaluation of results and discusses the findings. Finally, Section 6 concludes the paper with a summary of contributions and potential future research directions.

%\section{Related Work on Residual Networks in Engineering}
%In related work, Lu et al. \cite{LuLu20} leveraged recent neural network progress via a multifidelity (MFNN) strategy (MFNN: refers to a neural network architecture that combines outputs from multiple models with varying levels of fidelity or accuracy) for extracting material properties from instrumented indentation (see \cite{LuLu20}, Fig.~1(D) ).
% The proposed MFNN in this study incorporates a residual link (red line) connecting the low-fidelity output to the high-fidelity output.
%
%Wang et al. \cite{Wang21} proposed an improved fully-connected neural architecture. The key innovation involves integrating two transformer networks to project input variables into a high-dimensional feature space. This architecture combines multiplicative interactions and residuals, resulting in improved predictive accuracy, but with the cost of CPU time.

%\begin{figure}[!h]
%\centering%
%\includegraphics[width=6.3in]{ref_nets}
%\caption{Residual networks from Literature.} \label{ref_nets}
%\end{figure}

%\cite{Güler19}
\section{Plain Neural Networks}
%%%%%%%%%%%%%%%%%%%%%%%%%%%%
%%%%%%%%%%%%%%%%%%%%%%%%%%%%
%%%%%%%%%%%%%%%%%%%%%%%%%%%%%%
\subsection{Multi-Layer Perceptrons}

The objective of this section is to delve into the architecture and operations of Multi-Layer Perceptrons (MLPs), which form the foundation of many deep learning models. MLPs network, denoted as $\mathcal{F}$, approximate a function \(\textbf{u} : \textbf{x} \in \mathbb{R}^d \rightarrow \textbf{y} \in \mathbb{R}^D\) by stacking computing units called artificial neurons in consecutive layers. The structure of an MLP typically includes:

\begin{itemize}
    \item Source Layer: The zeroth layer  is called the source layer, responsible for providing an input (of dimension \(d\)) to the network.
    \item Hidden Layers: Every layer between the source and output layers is a hidden layer, where computations occur.
    \item Output Layer: The last layer  is the output layer, which produces the network’s prediction (of dimension \(D\)).
\end{itemize}

\begin{figure}
\centering%
\includegraphics[width=3.3in]{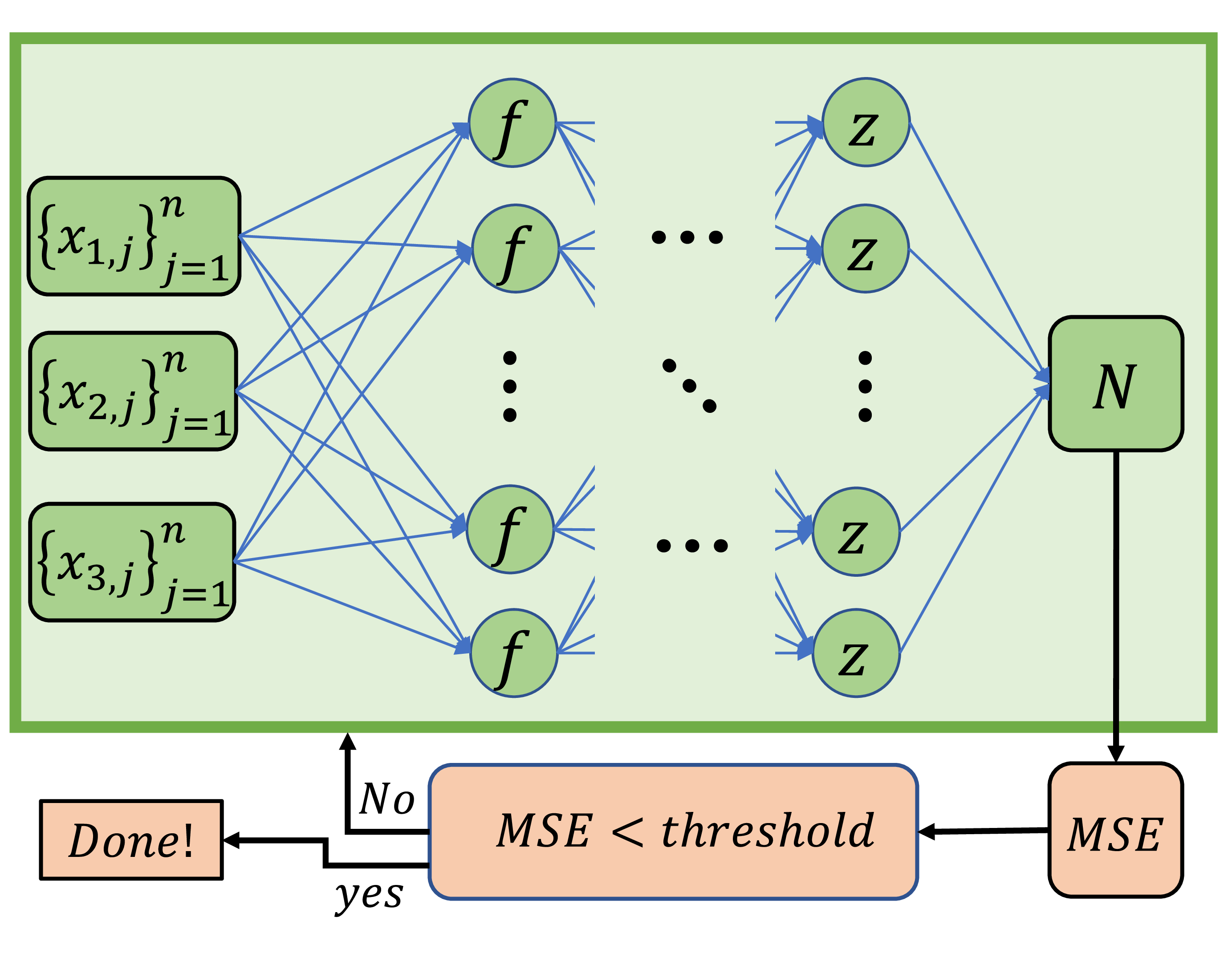}
\caption{schematic of an MLP.  } \label{MLP}
\end{figure}

The width of each layer, denoted as \(h^{(l)}\), determines the number of neurons in that layer.
We consider a network with \(L\) hidden layers, where the output vector for the \(l\)-th layer is denoted as \(\textbf{x}^{(l)} \in \mathbb{R}^{h^{(l)}}\), serving as the input to the next layer. The input signal provided by the input layer is denoted as \(\textbf{x}^{(0)} = \textbf{x} \in \mathbb{R}^d\), where $\textbf{x}=(x_1,x_2,\cdots,x_d)$.
In each layer \(l\), \(1 \leq l \leq L + 1\), the \(i\)-th neuron performs an affine transformation followed by a non-linear transformation:

\begin{equation}\label{eq1}
    z^{(l)}_i = W^{(l)}_{ij} {\rm{x}}^{(l-1)}_j + b^{(l)}_i, \quad 1 \leq i \leq h^{(l)}, \quad 1 \leq j \leq h^{(l-1)},
\end{equation}

\begin{equation}\label{eq2}
    {\rm{x}}^{(l)}_i = \sigma(z^{(l)}_i), \quad 1 \leq i \leq h^{(l)}.
\end{equation}
Here, \(W^{(l)}_{ij}\) and \(b^{(l)}_i\) represent the weights and biases associated with the \(i\)-th neuron of layer \(l\), respectively, while \(\sigma(\cdot)\) denotes the activation function, which, in our case, is $\tanh$. The overall behavior of the network, denoted as \(\mathcal{F} : \mathbb{R}^d \rightarrow \mathbb{R}^D\), can be conceptually understood as a sequence of alternating affine transformations and component-wise activations, as depicted in Eqs. \eqref{eq1}-\eqref{eq2}. The architecture of a Multilayer Perceptron is illustrated schematically in Fig. \ref{MLP}, where \(x_1\), \(x_2\), and \(x_3\) represent three dimensions ($\mathbf{x} = (x_1, x_2, x_3)$), each containing \(n\) samples. In this figure, \(\textbf{N}\) represents the approximated function, \(f\) embodies a fusion of linear (affine) and non-linear transformations of a plain neural network, expressed as:
\[
f(\textbf{x}) = \sigma(\textbf{Wx + b}),
\]
and \[
\textbf{z = Wx + b},
\] signifies the linear (affine) transformation.

\subsection{Network Parameters}

The parameters of the network consist of all the weights and biases, which we represent as follows:

\begin{itemize}
    \item We denote the parameters as \( \theta = \{\textbf{W}^{(l)}, \textbf{b}^{(l)}\}_{l=1}^{L+1}\).
    \item Each layer \(l\) has its weight matrix \(\textbf{W}^{(l)}\) and bias vector \(\textbf{b}^{(l)}\).
\end{itemize}
Therefore, the network $\mathcal{F}(\textbf{x}; \theta)$ represents a family of parameterized functions, where \(\theta\) needs to be suitably chosen such that the network approximates the target function \(\textbf{u}(\textbf{x})\) at the input \(\textbf{x}\).

%
%\subsection{Loss Function}
%
%In the context of training neural networks, selecting an appropriate loss function is pivotal. One commonly utilized measure is the mean squared error (MSE), defined as:
%
%\begin{equation}
%MSE = \frac{1}{n} \sum^{n}_{j=1} (\textbf{u}({\textbf {x}}_j) - \textbf{N}({\textbf {x}}_j))^2
%\end{equation}
%where predicted values, denoted as $\textbf{N}$, and the actual data values, denoted as $\textbf{u}$, are compared at individual data points identified as ${{\textbf {x}}_j}$. The $MSE$ is computed across a total of $n$ data points. This loss function, known as the $MSE$, guides the optimization process by directing the network towards generating accurate predictions.

\subsection{Training, Validation, and Testing of Neural Networks}
In the realm of supervised learning, training,  and testing are essential phases for optimizing and evaluating neural networks.
Let \( \mathcal{S} = \{(\textbf{x}_i, {\textbf{y}}_i) : 1 \leq i \leq n\} \) represent a dataset of pairwise samples corresponding to a target function \( \textbf{u} : \textbf{x} \rightarrow \textbf{y} \), {where $\textbf{x}$ represents the input data (coordinates) and $\textbf{y}$ represents the ground truth (also called the right-hand side in classical methods).Physics-Informed with Power-Enhanced Residual Network for Function Approximation and Inverse Problems} The objective is to approximate this function using the neural network \( \mathcal{F}(\textbf{x}; \theta, \phi) \), where \( \theta \) represents network parameters and \( \phi \) denotes hyperparameters such as depth, width, and activation function type.
The network optimization process involves two primary steps:

\begin{enumerate}
    \item \textbf{Training Phase:}
Training the neural network involves utilizing the training set \( \mathcal{S}_{\text{train}} \) over  \( n_{\text{train}} \) train points, and denoted as \( \mathcal{L}_{\text{test}} \),  to address the following optimization problem: Determine the optimal parameters \( \theta^* \) by minimizing the training loss function \( \mathcal{L}_{\text{train}}(\theta) \), defined as:

\begin{equation}
    \theta^* = \arg \min_\theta \mathcal{L}_{\text{train}}(\theta),
\end{equation}

\begin{equation}
    \mathcal{L}_{\text{train}}(\theta) = \frac{1}{n_{\text{train}}} \sum_{\substack{i=1, \\ (\mathbf{x}_i,\mathbf{y}_i) \in \mathcal{S}_{\text{train}}}}^{n_{\text{train}}}
 \| \mathbf{y}_i - \mathcal{F}(\mathbf{x}_i; \theta, {\phi}) \|_2^2.
\end{equation}
where \( {\phi} \) denotes a fixed set of hyperparameters {such as the learning rate, the number of hidden layers, the number of neurons in each hidden layer, the optimization methods used, etc} The optimal \( \theta^* \) is determined using an appropriate gradient-based algorithm (to be discussed in Section 5). This loss function \( \mathcal{L}_{\text{train}} \) is commonly referred to as the mean-squared loss function.

    \item \textbf{Validation or Testing Phase:}
After deriving the ``optimal'' network defined by \( \theta^* \) and \( \phi^* \), it undergoes evaluation using the test set \( \mathcal{S}_{\text{test}} \) to assess its performance on previously unseen data, thereby validating its efficacy beyond the initial training phases.

\end{enumerate}
To facilitate these phases, the dataset \( \mathcal{S} \) is typically split into training, and test sets, ensuring that the network's performance is rigorously evaluated across various datasets.

%%%%%%%%%%%%%%%%%%%%%%%%%%%%%%%
%%%%%%%%%%%%%%%%%%%%%%%%%%%%%%
%%%%%%%%%%%%%%%%%%%%%%%%%%%%%%%
\subsection{Calculating Gradients using Back-propagation}

In this section, we focus on understanding gradient evaluation during neural network training, specifically centering on backpropagation. Additional details can be found in \cite{Schmidhuber15}. An essential aspect of the training algorithm is gradient evaluation during network training. Recall the expression for the output \( \textbf{x}^{(l+1)} \) at the \( l + 1 \) layer:\\
     \textbf{Affine transformation:}
    \begin{equation}\label{zeq}
        z^{(l+1)}_i = W^{(l+1)}_{ij} x^{(l)}_j + b^{(l+1)}_i, \quad 1 \leq i \leq h^{(l+1)}, \quad 1 \leq j \leq h^{(l)}.
    \end{equation}
     \textbf{Non-linear transformation:}
    \begin{equation}\label{sigmaeq}
        x^{(l+1)}_i = \sigma\left(z^{(l+1)}_i\right), \quad 1 \leq i \leq h^{(l+1)}.
    \end{equation}
Considering a training instance denoted as \(\textbf{(x, y)}\), let \(\textbf{x}^{(0)} = \textbf{x}\). The loss function value  can be assessed through  the forward pass:
\begin{itemize}
    \item[(i)] Alpha For \( l = 1, \ldots, L + 1 \),
    \begin{itemize}
        \item[(a)] Find \( \textbf{z}^{(l)} \) from \eqref{zeq}.
        \item[(b)] Find \( \textbf{x}^{(l)} \) from \eqref{sigmaeq}.
    \end{itemize}
    \item[(ii)] Alpha Assess the loss function as:
    \[
    {\mathcal{L}}(\theta) = \|\textbf{y} - {\mathcal{F}}({\textbf{x}}; \theta, \phi)\|^2.
    \]
\end{itemize}
To update the network parameters, the derivatives \( \frac{\partial {\mathcal{L}}}{\partial \theta} \), or more specifically \( \frac{\partial {\mathcal{L}}}{\partial \textbf{W}^{(l)}}, \frac{\partial {\mathcal{L}}}{\partial \textbf{b}^{(l)}} \) for \( 1 \leq l \leq L + 1 \) are required. {The following algorithm outlines the necessary steps:}

\begin{enumerate}
    \item Expressions for these derivatives are derived by initially obtaining expressions for \( \frac{\partial {\mathcal{L}}}{\partial {\textbf{z}}^{(l)}} \) and \( \frac{\partial {\mathcal{L}}}{\partial \textbf{x}^{(l)}} \). Applying the chain rule iteratively results in:
    \begin{equation}\label{chain}
    \frac{\partial {\mathcal{L}}}{\partial {\textbf{z}}^{(l)}} = \frac{\partial {\mathcal{L}}}{\partial \textbf{x}^{(L+1)}} \cdot \frac{\partial \textbf{x}^{(L+1)}}{\partial {\textbf{z}}^{(L+1)}} \cdot \frac{\partial {\textbf{z}}^{(L+1)}}{\partial \textbf{x}^{(L)}} \cdots \frac{\partial \textbf{x}^{(l+1)}}{\partial {\textbf{z}}^{(l+1)}} \cdot \frac{\partial {\textbf{z}}^{(l+1)}}{\partial \textbf{x}^{(l)}} \cdot \frac{\partial \textbf{x}^{(l)}}{\partial {\textbf{z}}^{(l)}}    .
    \end{equation}

    \item To evaluate this expression, the following terms need to be computed:
    \begin{equation}
        \frac{\partial {\mathcal{L}}}{\partial \textbf{x}^{(L+1)}} = -2(\textbf{y} - \textbf{x}^{(L+1)})^T,
    \end{equation}
    and
    \begin{equation}
        \frac{\partial {\textbf{z}}^{(l+1)}}{\partial \textbf{x}^{(l)}} = \textbf{W}^{(l+1)},
    \end{equation}
    and
    \begin{equation}
        \frac{\partial \textbf{x}^{(l)}}{\partial {\textbf{z}}^{(l)}} = \textbf{M}^{(l)} \equiv \text{diag}[\sigma'( {\textbf{z}}^{(l)}_1 ), \ldots , \sigma'( {\textbf{z}}^{(l)}_{h^{(l)}} )].
    \end{equation}
    \item Using relations in \eqref{chain}, we obtain:
    \begin{equation}
    \frac{\partial {\mathcal{L}}}{\partial {\textbf{z}}^{(l)}} = \textbf{M}^{(l)} \textbf{W}^{(l+1)T} \textbf{M}^{(l+1)} \cdots \textbf{W}^{(L+1)T} \textbf{M}^{(L+1)} [-2(\textbf{y} - \textbf{x}^{(L+1)})].
    \end{equation}

    \item The final step is to derive an explicit expression for \( \frac{\partial {\mathcal{L}}}{\partial \textbf{W}^{(l)}} \). This can be accomplished by recognizing:
\begin{equation}\label{bp}
    \frac{\partial {\mathcal{L}}}{\partial \textbf{W}^{(l)}} = \frac{\partial {\mathcal{L}}}{\partial {\textbf{z}}^{(l)}} \cdot \frac{\partial {\rm z}^{(l)}}{\partial \textbf{W}^{(l)}} = \frac{\partial {\mathcal{L}}}{\partial {\textbf{z}}^{(l)}} \otimes \textbf{x}^{(l-1)}.
\end{equation}
 Here \( [\textbf{x} \otimes \textbf{y}]_{ij} = {\rm x}_i {\rm y}_j \) represents the outer product. Therefore, to evaluate \( \frac{\partial {\mathcal{L}}}{\partial \textbf{W}^{(l)}} \), both \( \textbf{x}^{(l-1)} \), evaluated during the forward phase, and \( \frac{\partial {\mathcal{L}}}{\partial {\textbf{z}}^{(l)}} \), evaluated during back-propagation, are required.
\end{enumerate}

\section{ Physics-Informed Neural Network for Solving Inverse Burgers' Equation}

\begin{figure}
\centering%
\includegraphics[width=5.3in]{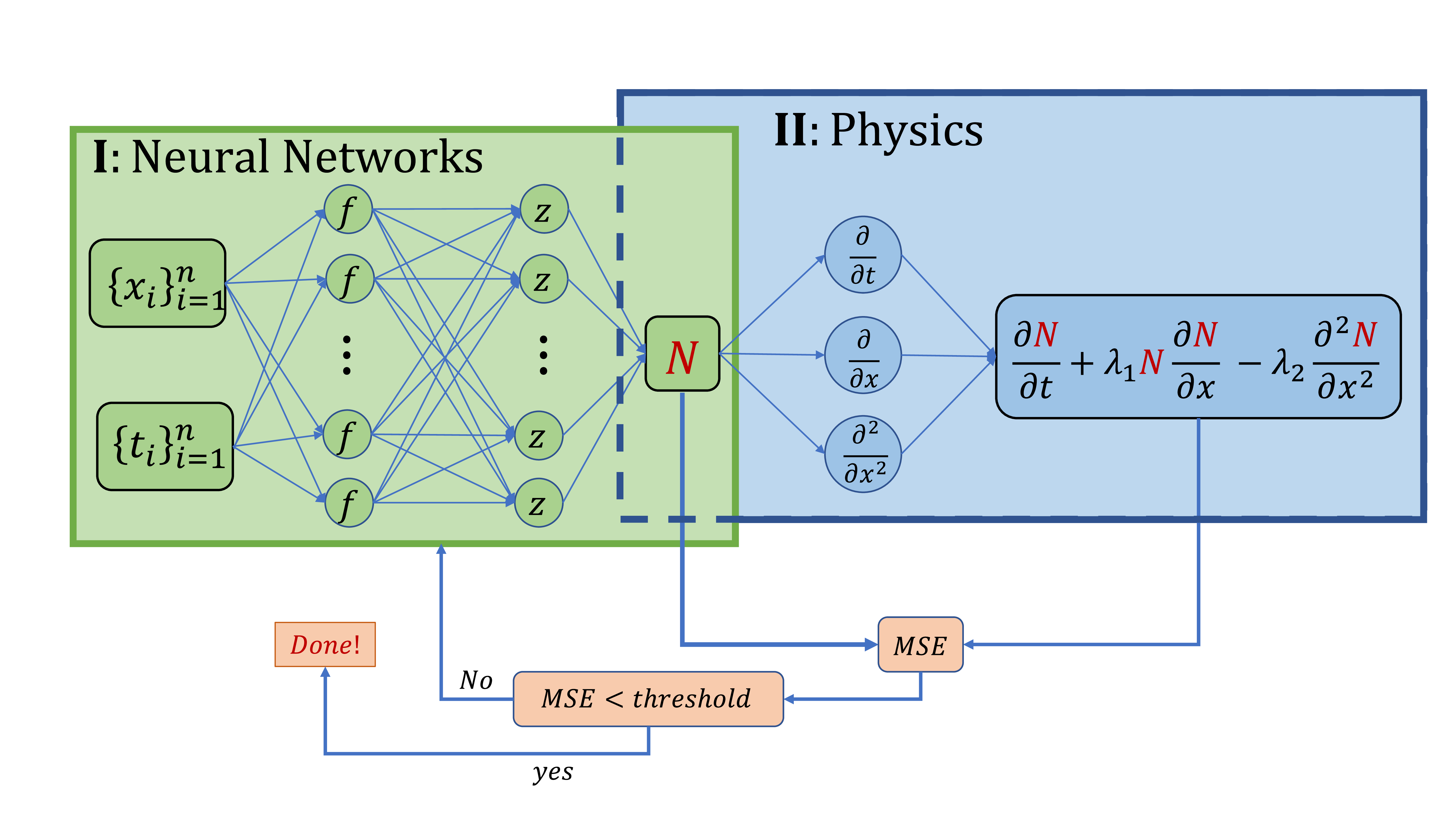}
\caption{The neural network (interpolation stage) + physics (inverse Burger's equation). Here, $x$ and $t$ represent two dimensions, each including $n$ samples. } \label{pinn}
\end{figure}

In this section, we explore the application of Physics-Informed Neural Networks \cite{Raissi19} to solve the inverse Burgers' equation in one dimension. The 1D Burgers' equation is given by:
\begin{equation}
\frac{\partial \textbf{u}}{\partial \textbf{t}} + \lambda_1 \textbf{u} \frac{\partial \textbf{u}}{\partial \textbf{x}} = \lambda_2 \frac{\partial^2 \textbf{u}}{\partial \textbf{x}^2},
\end{equation}
where \(\textbf{u}(\textbf{x},\textbf{t})\) is the solution, and \(\lambda_1\) and \(\lambda_2\) are coefficients to be determined. Here, \(\textbf{x} \in [-1, 1]\) and \(\textbf{t} \in [0, 1]\) represent two dimensions, space and time respectively.
In the context of solving the inverse Burgers' equation, we combine the power of neural networks (Fig.~\ref{pinn}(I)) with the physical governing equation (Fig.~\ref{pinn}(II)) to form PINN. Utilizing the universal approximation theorem, we approximate the solution \(\textbf{N}(\textbf{x}, \textbf{t}) \approx \textbf{u}(\textbf{x}, \textbf{t})\). By automatically differentiating the network, we can compute derivatives such as \(\textbf{N}_\textbf{t}=\frac{\partial \textbf{N}}{\partial \textbf{t}}\), \(\textbf{N}_{\textbf{xx}}=\frac{\partial^2 \textbf{N}}{\partial \textbf{x}^2}\), etc.
We define the function \(\textbf{g}(\textbf{x},\textbf{t})\) representing the residual of the Burgers' equation as (Fig.~\ref{pinn}(II)):
\begin{equation}
\textbf{g}(\textbf{x},\textbf{t}) = \textbf{N}_\textbf{t} + \lambda_1 \textbf{N} \textbf{N}_\textbf{x} - \lambda_2 \textbf{N}_{\textbf{xx}}.
\end{equation}
The PINNs loss function is given by :
\begin{equation}
MSE_\textbf{g} = \frac{1}{n_c} \sum^{n_c}_{i=1} (\textbf{g}({\rm x}^i,{\rm t}^i))^2.
\end{equation}
Here $\textbf{x}$ and $\textbf{t}$ respectively represent the spatial and temporal coordinates for the Burgers' equation. The superscript $i$ denotes the index of the collocation points where $1\leq i \leq n_c$, with $n_c$ denoting the number of collocation points. It is important to note that $n_c$ refers to the number of data points involved in the PDE loss calculation, which may differ from $n$, representing the number of observed data in the context of inverse problem (see \cite{Raissi19}). In this inverse problem scenario, we incorporate $n$ observed data to compute the loss with respect to the reference solution, as shown in Figure \ref{pinn}(I):
\begin{equation}
MSE_\textbf{u} = \frac{1}{n} \sum^{n}_{j=1} \left(\textbf{u}({\rm x}^i,{\rm t}^i) - \textbf{N}({\rm x}^j,{\rm t}^j)\right)^2,
\end{equation}
where \(\textbf{N}\) represents the solution provided by the network, as depicted in Fig.~\ref{pinn}(I), and \(\textbf{u}\) denotes the reference solution both over \(n\) samples.
 The total loss function minimized during training is:
\begin{equation}
MSE = MSE_{\textbf{u}} + MSE_\textbf{g}.
\end{equation}
  We aim to minimize \(MSE\) to obtain the neural network parameters \( \theta = \{\textbf{W}^{(l)}, \textbf{b}^{(l)}\}_{l=1}^{L+1}\) in each layer \(l\), \(1 \leq l \leq L + 1\), and the Burgers' equation parameters \(\lambda_1\) and \(\lambda_2\).

\section{Advanced Deep Neural Network Architectures}
\subsection{Highway Networks}
This section introduces Highway Networks, a variant of deep neural networks designed to facilitate effective information flow across multiple layers \cite{Srivastava15a,Srivastava15}. Highway Networks incorporate gating mechanisms that regulate information flow, allowing selective transformation and retention of input information.
A plain neural network can be expressed as:
\begin{equation}
\textbf{x}^{(l)} = f^{(l)}_{pn}(\textbf{x}^{(l-1)}),
\end{equation}
where $\textbf{x}$ represents the input to the layer and $f^{(l)}_{pn}(\textbf{x}^{(l-1)})$ encompasses the sequence of operations (\eqref{eq1} and \eqref{eq2}) at layer $l-1$.
Highway Networks integrate two supplementary gating mechanisms responsible for:
\begin{itemize}
    \item Applying typically non-linear transformations (controlled by the transform gate \( T \)),
    \item Deciding how much activation from the previous layer gets copied to the current layer (controlled by the carry gate \( C \)).
\end{itemize}
This is illustrated by the equation:
\begin{equation}\label{hn}
\textbf{x}^{(l)} = f^{(l)}_{hw}(\textbf{x}^{(l-1)})\cdot T^{(l)}(\textbf{x}^{(l-1)}) + \textbf{x}^{(l-1)}\cdot C^{(l)}(\textbf{x}^{(l-1)}),
\end{equation}
where $f^{(l)}_{hw}(\textbf{x}^{(l-1)})$ represents the sequence of the linear and non-linear functions for the highway neural network. This formulation allows Highway Networks to learn both feedforward and shortcut connections \cite{He15, He16} simultaneously, enabling effective training of very deep networks while retaining valuable information throughout the network's layers. Highway Networks have demonstrated promising results in various tasks, including image classification, speech recognition, and natural language processing, showcasing their efficacy in facilitating the training of deep neural networks \cite{Srivastava15a,Srivastava15}.

\subsection{Residual Network}

 Residual Networks offer a streamlined approach compared to Highway Networks by redefining the desired transformation as the input augmented by a residual. Residual networks, often denoted as ResNets \cite{He15, He16}, have emerged as a prominent architecture in neural networks, representing a specialized case of Highway networks where both $C$ and $T$ are set to 1 and remain fixed {in} \eqref{hn} \cite{Greff17}. They are characterized by their residual modules, denoted as $f^{(l)}_{rs}$, and skip connections that bypass these modules, enabling the construction of deep networks. This allows for the creation of residual blocks, which are sets of layers within the network. In contrast with Fig.~\ref{ExNet}(a), which illustrates the plain neural network, Fig.~\ref{ExNet}(b) showcases the network architecture incorporating ResNet features. To simplify notation, the initial pre-processing and final  steps are excluded from our discussion. Therefore, the definition of the output $\textbf{x}^{(l)}$    for the $l$-th layer is given as follows:
\begin{equation}
\textbf{x}^{(l)} = f^{(l)}_{rs}(\textbf{x}^{(l-1)}) + \textbf{x}^{(l-1)}, \label{eq:residual}
\end{equation}
where $f^{(l)}_{rs}(\textbf{x}^{(l-1)})$ encompasses a sequence of operations, including \textit{linear transformations} \eqref{eq1}, \textit{element-wise activation functions} \eqref{eq2} at layer $l-1$ with $1\leq l \leq L$ for residual network.
\subsection{Proposed SQR-SkipResNet}
\begin{figure}[!h]
\centering%
\includegraphics[width=6.3in]{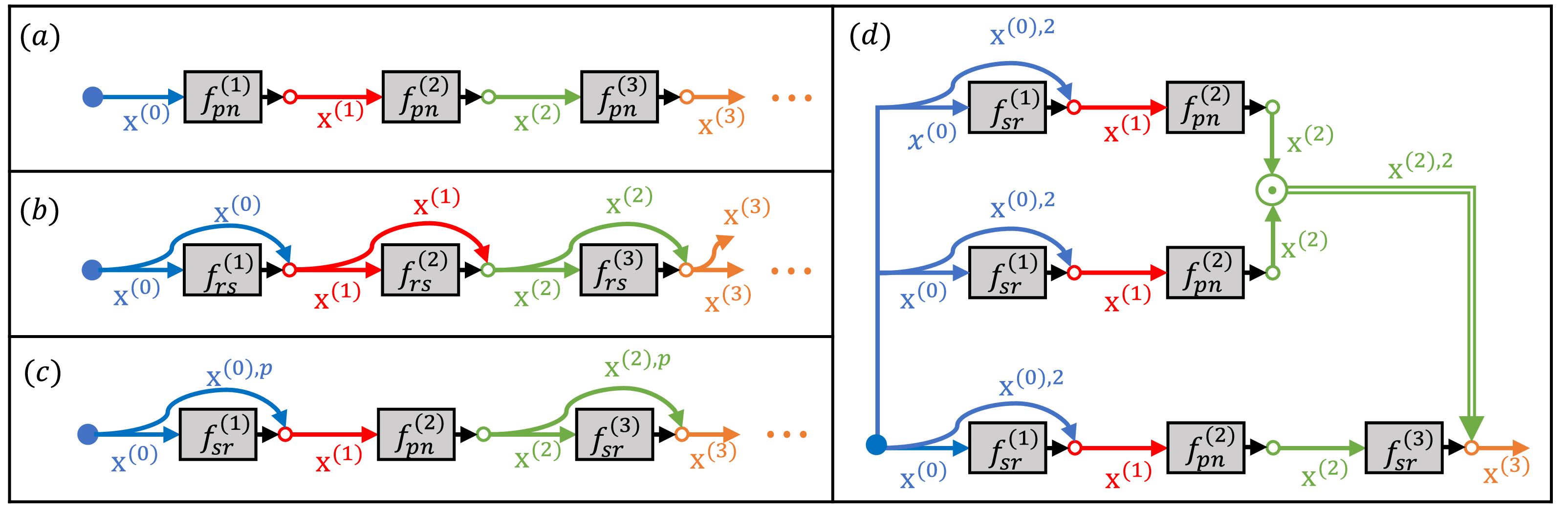}
\caption{Three neural network architectures: (a) plain neural network (Plain NN), (b) residual network (ResNet), (c) power-enhanced   SkipResNet, and (d) Unraveled SQR-SkipResNet (plot (c) with $p=2$) where $\odot$ denotes element-wise multiplication.} \label{ExNet}
\end{figure}

In this study, we propose a power-enhanced variant of the ResNet that skips every other layer, denoted as the ``SQR-SkipSkipResNet.'' The modification involves altering the recursive definition in \eqref{eq:residual} as follows:
\begin{equation}
\left\{
\begin{aligned}
    \textbf{x}^{(l)} &= f^{(l)}_{sr}(\textbf{x}^{(l-1)}) + \textbf{x}^{{(l-1)},p}, & \text{for}\quad l = 1,3,5,\ldots\\
    \textbf{x}^{(l)} &= f_{pn}^{(l)}(\textbf{x}^{(l-1)}), & \text{for}\quad l = 2,4,6,\ldots
\end{aligned}
\right.
\end{equation}
where $f^{(l)}_{sr}(\textbf{x}^{(l-1)})$ denotes the sequence of linear and non-linear operations for the proposed SQR-SkipResNet. This novel configuration, illustrated in Fig.~\ref{ExNet}(c), introduces the use of a power term $\textbf{x}^{{(l-1)},p}$ for specific layers, enhancing the expressive power of the network.
%The subsequent sections will delve into the details of this enhanced architecture and its implications for performance improvement.

For the purpose of comparison among Plain NN, ResNet, and SQR-SkipResNet (Figs.~\ref{ExNet}(a)-(c), respectively), we evaluate the output of the third hidden layer concerning the input $\textbf{x}_0$. The results for the plain neural network are as follows:
\begin{flalign}
\textbf{x}^{(3)} &= f_{pn}^{(3)}(\textbf{x}^{(2)}) & \nonumber\\
&= f_{pn}^{(3)}(f_{pn}^{(2)}(\textbf{x}^{(1)})) & \nonumber\\
&= f_{pn}^{(3)}(f_{pn}^{(2)}(f_{pn}^{(1)}(\textbf{x}^{(0)}))). &
\end{flalign}
Meanwhile, the corresponding ResNet formulation is as follows \cite{Veit16}:
\begin{flalign}
\textbf{x}^{(3)} &= f_{rs}^{(3)}(\textbf{x}^{(2)}) + \textbf{x}^{(2)} & \nonumber \\
&= f_{rs}^{(3)}(f_{rs}^{(2)}(\textbf{x}^{(1)}) + \textbf{x}^{(1)}) + \left[f_{rs}^{(2)}(\textbf{x}^{(1)}) + \textbf{x}^{(1)}\right] &\nonumber \\
&= f_{rs}^{(3)}(f_{rs}^{(2)}(f_{rs}^{(1)}(\textbf{x}^{(0)}) + \textbf{x}^{(0)}) + f_{rs}^{(1)}(\textbf{x}^{(0)}) + \textbf{x}^{(0)}) + \left[f_{rs}^{(2)}(f_{rs}^{(1)}(\textbf{x}^{(0)}) + \textbf{x}^{(0)}) + f_{rs}^{(1)}(\textbf{x}^{(0)}) + \textbf{x}^{(0)} \right]. &
\end{flalign}
Finally, the formulation of the first three hidden layers for the SQR-SkipResNet is as follows:
\begin{flalign}
\textbf{x}^{(3)} &= f_{sr}^{(3)}(\textbf{x}^{(2)}) + \textbf{x}^{{(2)},p} &\nonumber \\
&= f_{sr}^{(3)}(f_{sr}^{(2)}(\textbf{x}^{(1)})) + \left[f_{sr}^{(2)}(\textbf{x}^{(1)})\right]^p &\nonumber \\
&= f_{sr}^{(3)}(f_{sr}^{(2)}(f_{sr}^{(1)}(\textbf{x}^{(0)}) + \textbf{x}^{{(0)},p}) ) + \left[f_{sr}^{(2)}(f_{sr}^{(1)}(\textbf{x}^{(0)}) + \textbf{x}^{{(0)},p})\right]^p.&
\end{flalign}
Figure~\ref{ExNet}(d) visually represents the ``expression tree'' for the case with $p=2$, providing an insightful illustration of the data flow from input to output. The graph demonstrates the existence of multiple paths that the data can traverse. Each of these paths represents a distinct configuration, determining which residual modules are entered and which ones are skipped.

\noindent
Our extensive numerical experiments support our approach, indicating that a power of 2 is effective for networks with fewer than 30 hidden layers. However, for deeper networks, a larger power can contribute to network stability. Nonetheless, deploying such deep networks does not substantially enhance accuracy and notably increases CPU time. In tasks like function approximation and solving PDEs, a power of 2 generally suffices, and going beyond may not justify the added complexity in terms of accuracy and efficiency.

\section{Numerical Results}\label{section:numexa}
In this study, we employ the notations \(n\), \(n_l\), and \(n_n\) to represent the number of data points (training), layers, and neurons in each layer, respectively. In all following examples, unless otherwise mentioned, we consider $100^2$
 validation data points. We also introduce three distinct types of error measurements between exact $\textbf{u}$ and approximated $\textbf{N}$ solutions:
\begin{enumerate}
    \item Mean Square Error: The training errors shown in the plotted graphs, relative to the iteration number, are computed using the mean square error criterion.\\
  \[ \text{Mean Square Error} = \frac{1}{n} \sum_{i=1}^{n} \left( {\textbf{u}}_i - {\textbf{N}}_{i} \right)^2 \].

    \item Relative L2 Norm Error: The validation errors, calculated over the test data and presented in the plotted graphs concerning the iteration number, are measured using the relative L2 norm error metric.\\
     \[ \text{Relative L2 Norm Error} = \frac{\| {\textbf{u}} - {\textbf{N}} \|_2}{\| {\textbf{u}} \|_2} \].

\item Maximum Absolute Error: When visualizing errors across the entire domain, whether in 2D or 3D scenarios, the error represented on the contour error plot is referred to as the maximum absolute error. It is important to note that the contour bars are scaled according to the largest error in the plot.\\
    \[ \text{Maximum Absolute Error} = \max \left| {\textbf{u}} - {\textbf{N}} \right| \].
\end{enumerate}
These error metrics provide valuable insights into the accuracy and convergence of the methods used in this study.
In this section four methods will be investigated.
\begin{enumerate}
\item {\tt Plain NN}: A conventional neural network without any additional modifications or residual connections (see Fig.~\ref{ExNet}(a)).

\item {\tt ResNet}: A residual neural network architecture where the output of each layer is obtained by adding the residual to the layer's output (see Fig.~\ref{ExNet}(b)).

\item {\tt SkipResNet}: An extension of ResNet, where the residual connection is applied every other layer, alternating between including and excluding the residual connection (see Fig.~\ref{ExNet}(c) where $p=1$).

\item {\tt SQR-SkipResNet}:  An innovative variation of the ResNet architecture, where the squared residual is added every other layer. In this approach, the output of each alternate layer is obtained by squaring the previous layer's output and adding the squared residual to it (see Fig.~\ref{ExNet}(c)-(d) where $p=2$).
\end{enumerate}

\noindent
In all our experiments, we primarily employ L-BFGS-B (Limited-memory Broyden-Fletcher-Goldfarb-Shanno with Box constraints) and occasionally, for comparison, we also use Adam (Adaptive Moment Estimation). Convergence, particularly with L-BFGS-B optimization, is identified by satisfying preset tolerance levels for gradient or function value change, or by reaching the defined maximum number of iterations, with a gradient tolerance of $1 \times 10^{-9}$, and a change in function value tolerance of $1 \times 10^{-9}$.

In the following section, we have several scenarios:
\begin{enumerate}
    \item We first examine the capability of our proposed algorithm using three 2D test functions:
    \begin{itemize}
        \item The first function is smooth,
        \item the second one exhibits a singularity outside the boundary,
        \item and the third one is non-smooth.
    \end{itemize}
    \item Next, we apply the interpolation to a real-case study: the interpolation of data from Mount Eden in New Zealand, which also involves a 2D interpolation problem.
    \item To further demonstrate efficiency, we extend our analysis to a 3D example using the Stanford Bunny dataset.
    \item Finally, to showcase the versatility of the proposed method, we tackle an inverse partial differential equation, specifically Burger's equation. Further investigations into solving PDEs will be carried out in future research.
\end{enumerate}

The numerical experiments were
executed on a computer equipped
with an Intel(R) Core(TM) i9-9900 CPU operating at
3.10GHz  with a
total of 64.0 GB of RAM.

%%%%%%%%%%%%%%%%%%%%%%%%%%%%%%%%
%%%%%%%%%%%%%%%%%%%%%%%%%%%%%%%%%%%%
%%%%%%%%%%%%%%%%%%%%%%%%%%%%%%%%
%Example 1
\begin{example}\rm{
For the first example, three test functions are investigated and depicted in Fig.~\ref{Ex0_1}. The top panel of Fig.~\ref{Ex0_1} displays the 3D surface plot of the test functions, while the bottom panel presents the corresponding contour plots.
F1 is a smooth function, originally introduced by Franke \cite{Franke82}, which has been extensively used for studying radial basis function (RBF) interpolation. On the other hand, F2 and F3 are non-smooth functions \cite{Amir22b}.

\begin{align*}
{{\rm F1}(x_1,x_2)} & =  \frac{3}{4} \exp \left[ \frac{-1}{4} \left( (9x_1-2)^2+(9x_2-2)^2 \right) \right]
 +  \frac{3}{4}\exp \left[\frac{-1}{49}(9x_1+1)^2-\frac{1}{10}(9x_2+1)^2\right] \\
 & + \frac{1}{2}\exp\left[\frac{-1}{4} \left( (9x_1-7)^2+(9x_2-3)^2\right) \right]
- \frac{1}{5}\exp\left[-(9x_1-4)^2-(9x_2-7)^2\right],\\
{{\rm F2}(x_1,x_2)} &= \frac{0.0025}{(x_1-1.01)^2+(x_2-1.01)^2},\\
{{\rm F3}(x_1,x_2)} &= \frac{1}{9}\left[64 - 81\left(\left| x_1 - \frac{1}{2} \right| + \left| x_2 - \frac{1}{2} \right|\right)\right] - \frac{1}{2}.
\end{align*}

\begin{figure}
\centering%
\subfigure[F1]{ \label{F1_exact_3D}
\includegraphics[width=1.95in]{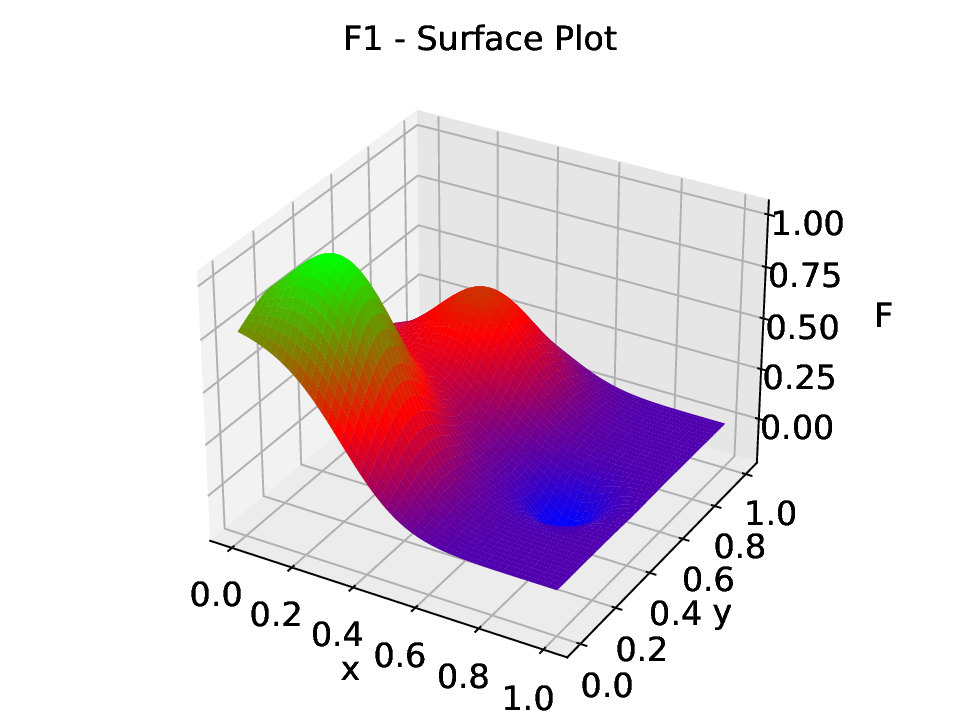}}
\subfigure[F2]{ \label{F2_exact_3D}
\includegraphics[width=1.95in]{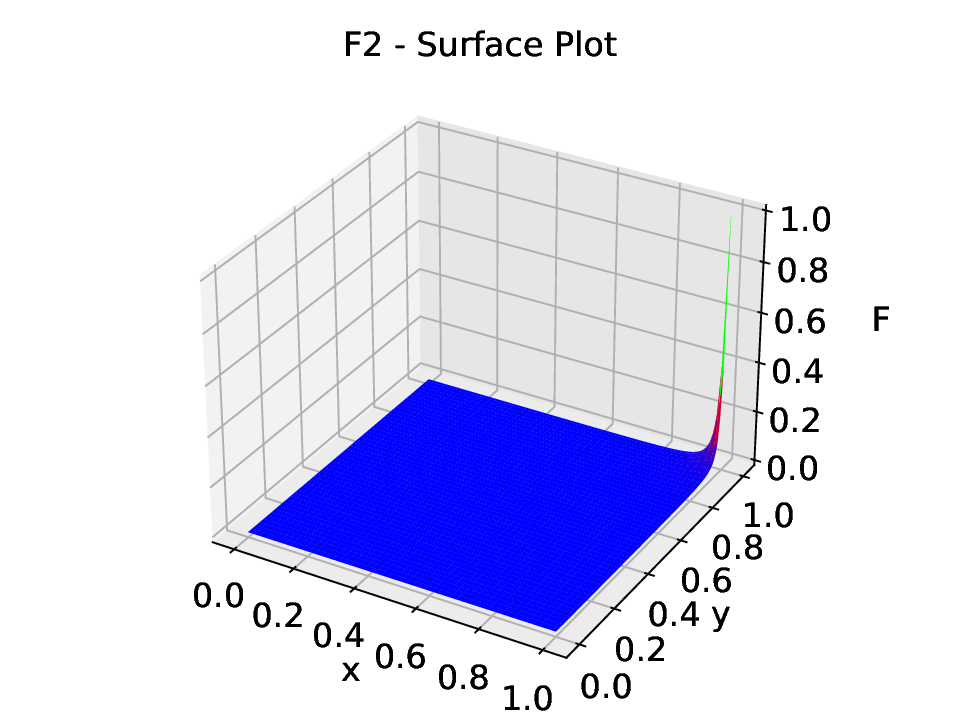}}
\subfigure[F3]{ \label{F3_exact_3D}
\includegraphics[width=1.95in]{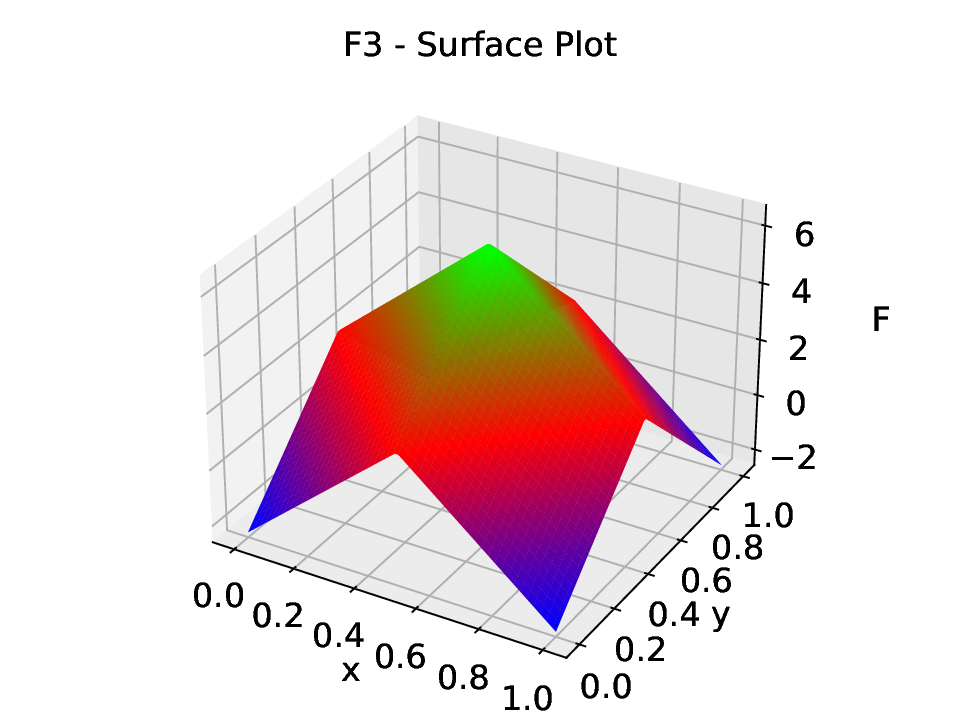}}
\subfigure[F1]{ \label{F1_exact_2D}
\includegraphics[width=1.95in]{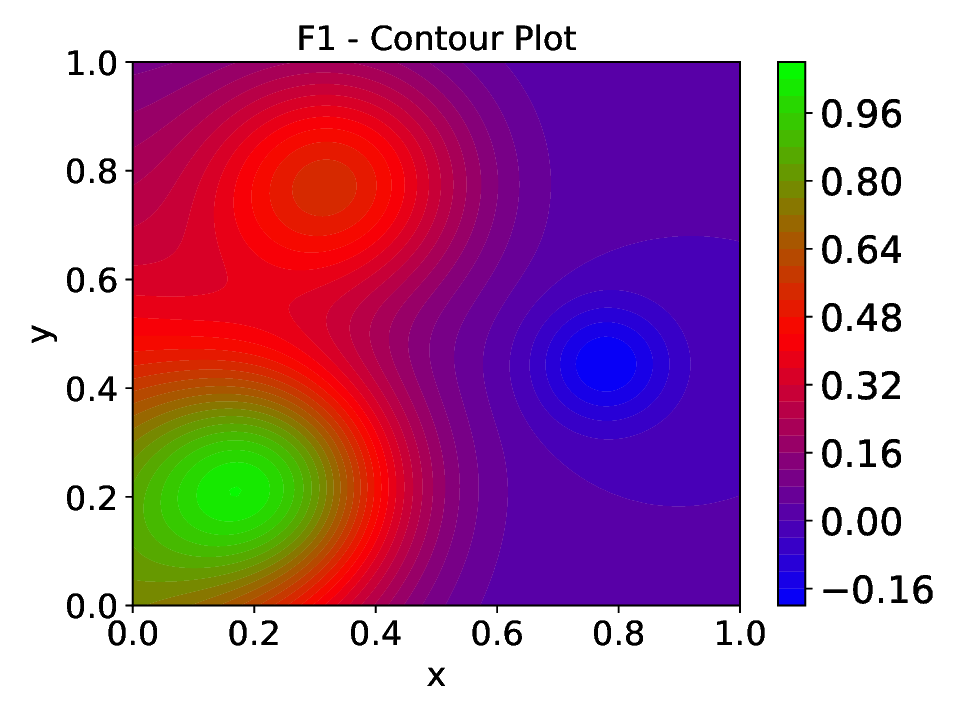}}
\subfigure[F2]{ \label{F2_exact_2D}
\includegraphics[width=1.95in]{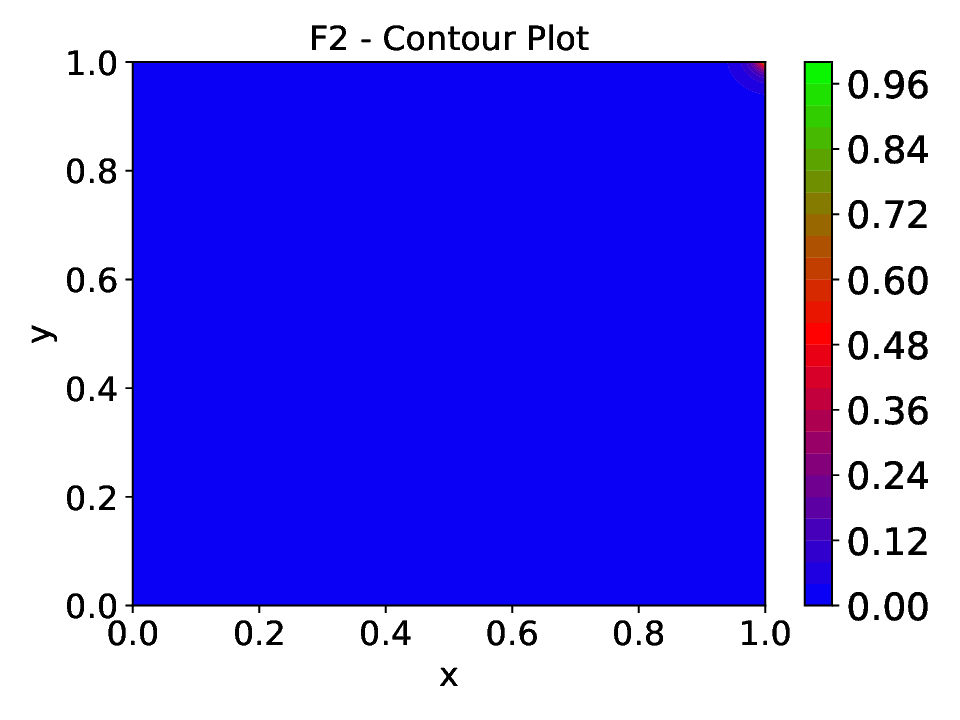}}
\subfigure[F3]{ \label{F3_exact_2D}
\includegraphics[width=1.95in]{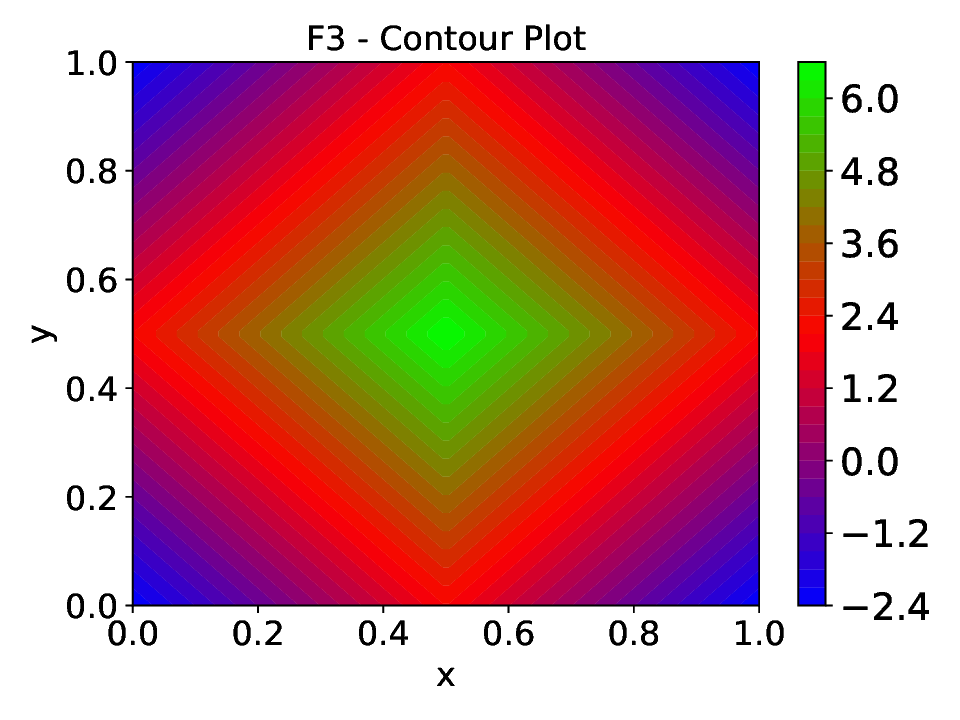}}
\caption{The profile of the F1, F2, and F3 } \label{Ex0_1}
\end{figure}

 \begin{figure}
\centering%
\subfigure[$n=500$]{ \label{Ex2_F1_n500_50x20}
\includegraphics[width=1.95in]{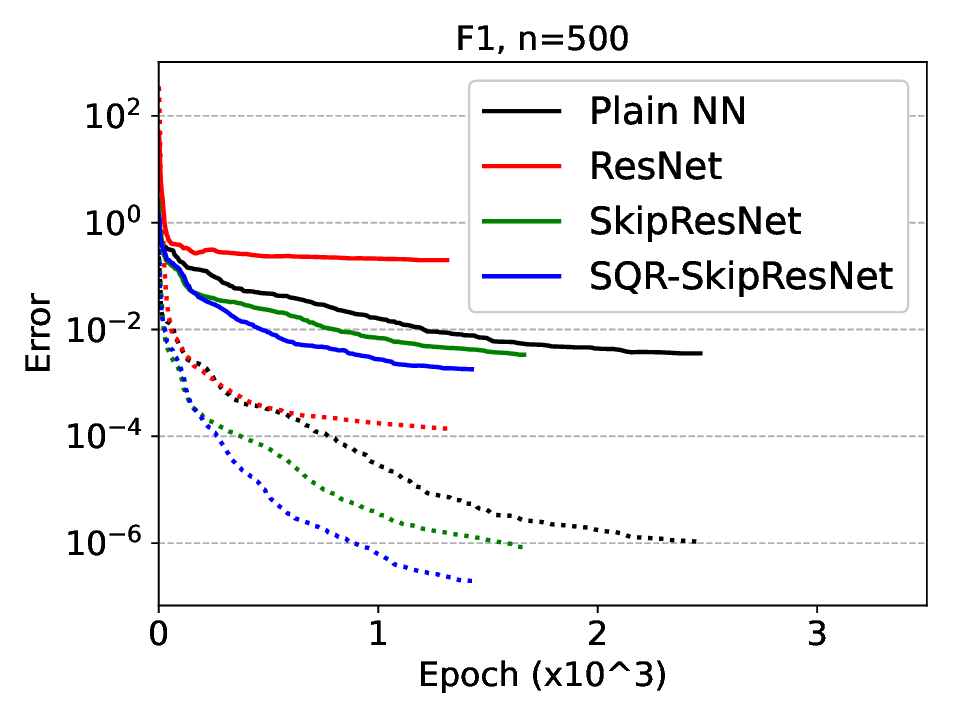}}
\subfigure[Plain NN]{ \label{Ex2_F1_n500_50x20_error_pnn}
\includegraphics[width=1.95in]{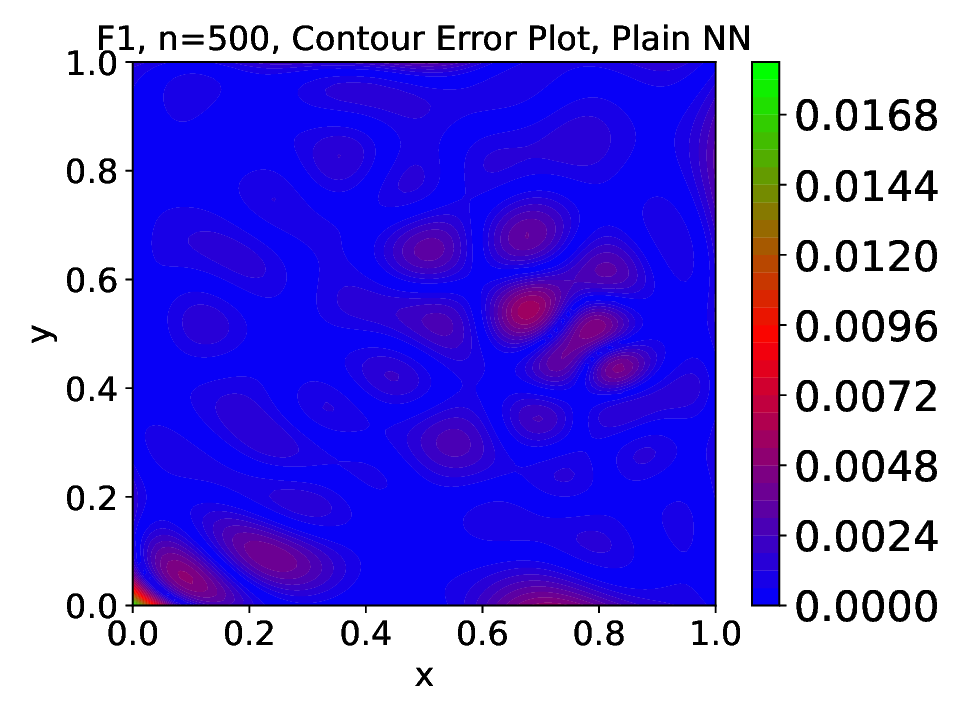}}
\subfigure[SQR-SkipResNet]{ \label{Ex2_F1_n500_50x20_error_sqr}
\includegraphics[width=1.95in]{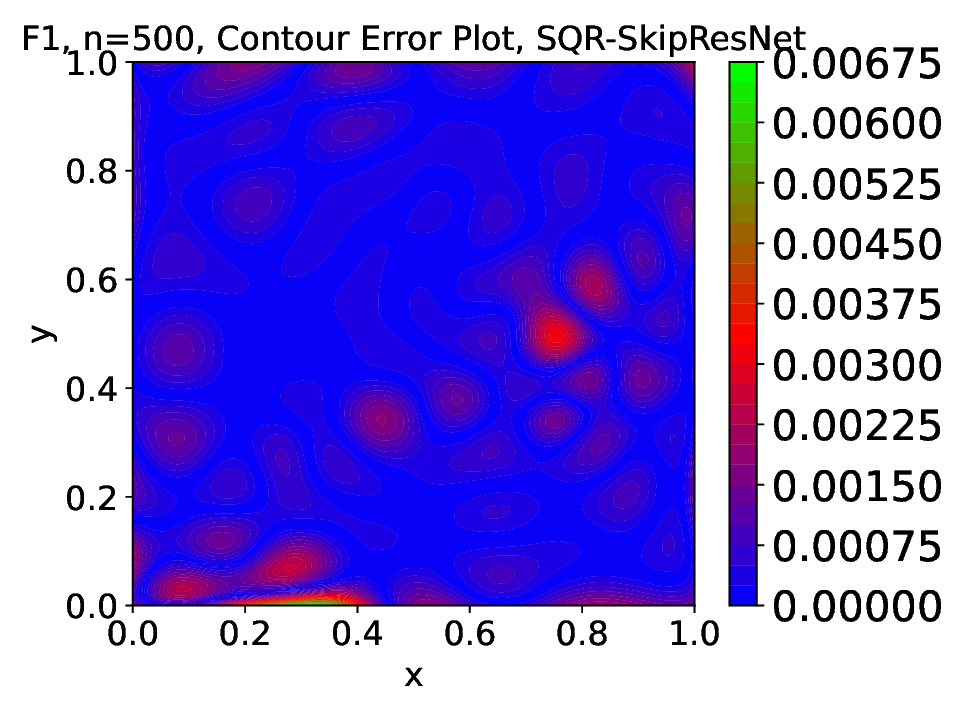}}
\caption{The profiles of training on F1 for different number of collocation points $n$. Dotted-line curves denote training error, and solid-line curves denote validation error. } \label{Ex1_1}
\end{figure}

First we investigate the performance of four neural networks: Plain NN, ResNet, SkipResNet, and SQR-SkipResNet for approximating of F1. The entire analysis is based on the network with $n_l=10$, and each layer contains 50 neurons ($n_n$).  {Figure~\ref{Ex1_1}} shows the results of interpolation using $500$ training data and $100^2$ validation data. Figure~\ref{Ex2_F1_n500_50x20} presents the Mean Squared Error over training (dashed line) and Relative L2 Norm over validation (solid line) data points. Figure~\ref{Ex2_F1_n500_50x20_error_pnn}-\ref{Ex2_F1_n500_50x20_error_sqr} show the maximum absolute errors for Plain NN and SQR-SkipResNet, respectively.

Our observations from these plots are as follows:

\begin{enumerate}
\item Plot (a) indicates that the ResNet is not accurate enough compared to the other three networks, both during training and validation. This pattern has been consistently observed in various examples, and we will no longer investigate the ResNet performance.

\item As indicated by the plot, it can be observed that the Plain NN necessitates approximately 2400 iterations for convergence, whereas the proposed SQR-SkipResNet achieves convergence in a significantly reduced 1400 iterations. Additionally, the latter method exhibits higher accuracy compared to the former.

\item Plot (a) also shows that SkipResNet performs somewhat between Plain NN and SQR-SkipResNet. This behavior has been observed in different examples conducted by the authors, but we do not plan to further investigate this method.

\item Contour error plots for both Plain NN and SQR-SkipResNet are presented in plots (b) and (c) respectively. These plots highlight that the maximum absolute error achieved with SQR-SkipResNet exhibits a remarkable improvement of approximately 60\% compared to Plain NN.
\end{enumerate}

Therefore, a higher accuracy and better convergence are observed when using SQR-SkipResNet compared to other algorithms.

\begin{figure}
\centering%
\subfigure[$n=5000$]{ \label{Ex2_F1_n5000_50x20}
\includegraphics[width=1.95in]{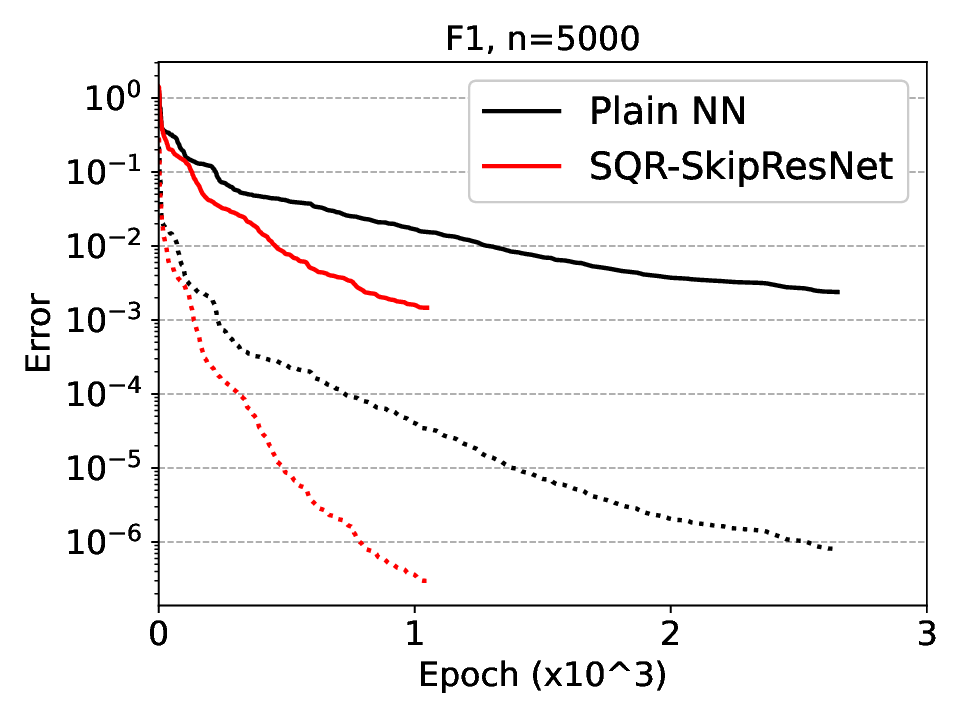}}
\subfigure[Plain NN]{ \label{Ex2_F1_n5000_50x20_err_pnn}
\includegraphics[width=1.95in]{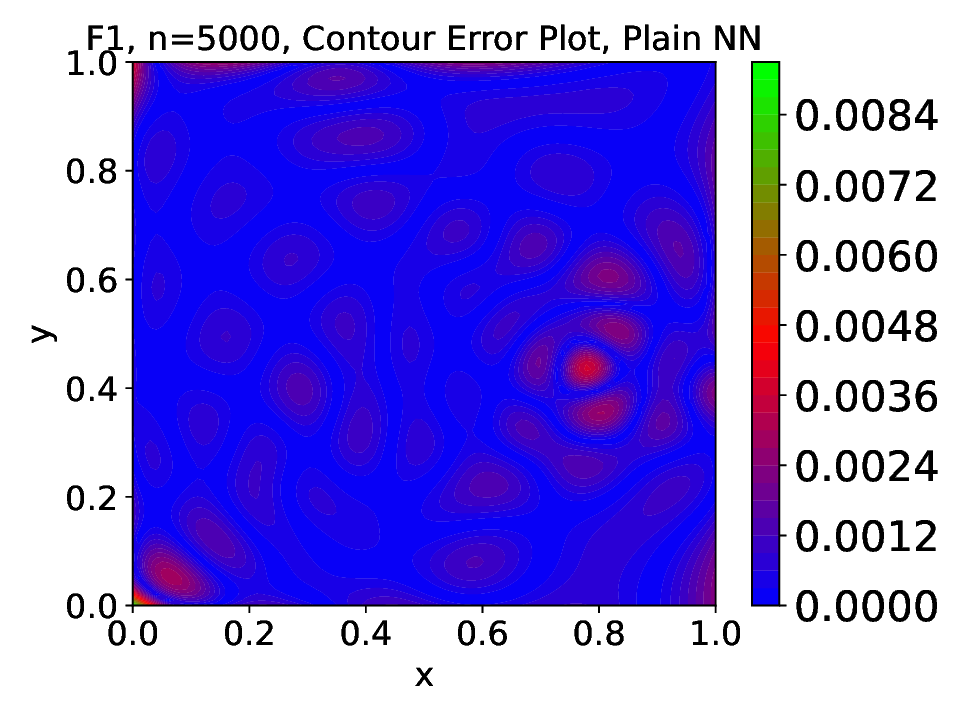}}
\subfigure[SQR-SkipResNet]{ \label{Ex2_F1_n5000_50x20_err_sqr}
\includegraphics[width=1.95in]{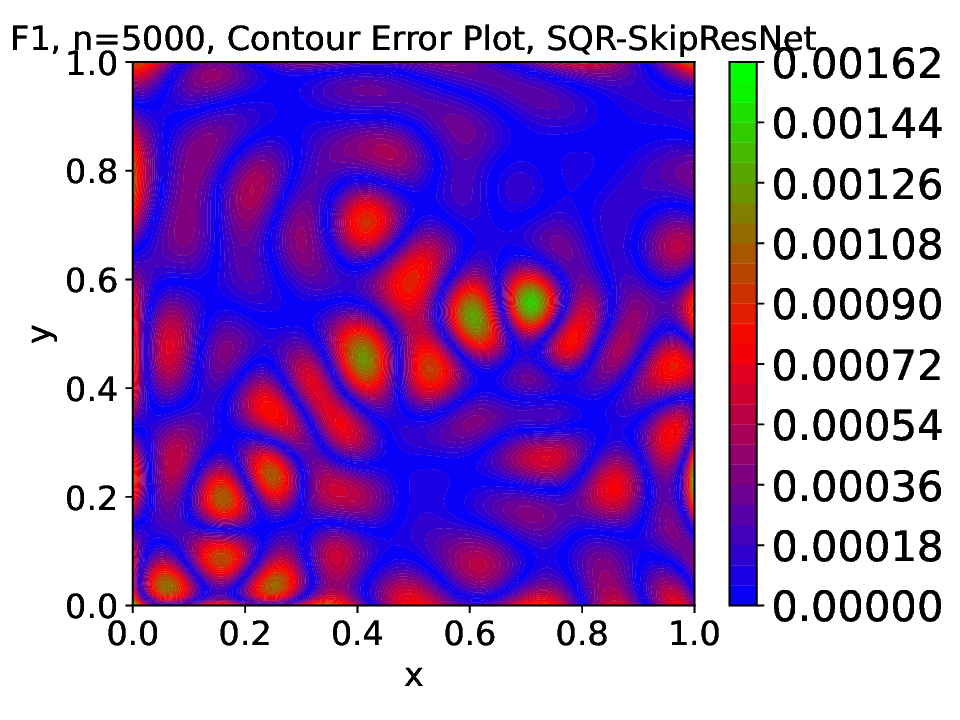}}
\caption{Example 1:  The profiles of (a) training and validation results on F1 with 5000 data points. Dotted-line curves denote training error, and solid-line curves denote validation error. The corresponding contour error plots for (b) the plain NN and (c) SQR-SkipResNet.} \label{Ex2_2}
\end{figure}

Fig.~\ref{Ex2_2} illustrates the outcomes obtained through the utilization of a large number of data points, $n=5000$, employed for the interpolation of F1.
 A better convergence from Fig.~\ref{Ex2_F1_n5000_50x20} can be observed using the proposed SQR-SkipResNet compared to Plain NN. The Plain NN yields a maximum absolute error of $9.07 \times 10^{-3}$ in 113 seconds, whereas the proposed SQR-SkipResNet approach achieves a significantly reduced error of $1.56 \times 10^{-3}$ in only 55 seconds, shown in Fig.~\ref{Ex2_F1_n5000_50x20_err_pnn}-\ref{Ex2_F1_n5000_50x20_err_sqr}, respectively. This represents an improvement of approximately $82.8\%$ in terms of error reduction and a substantial $51.3\%$ reduction in CPU processing time.
A comparison between Fig.~\ref{Ex2_F1_n500_50x20} and Fig.~\ref{Ex2_F1_n5000_50x20} reveals that a greater number of data values results in an improved convergence rate for the proposed SQR-SkipResNet, whereas the Plain NN exhibits a slightly higher iteration number.

\begin{figure}
\centering%
\subfigure[Plain NN]{ \label{F1_n500_50x10_normW_pnn}
\includegraphics[width=2.95in]{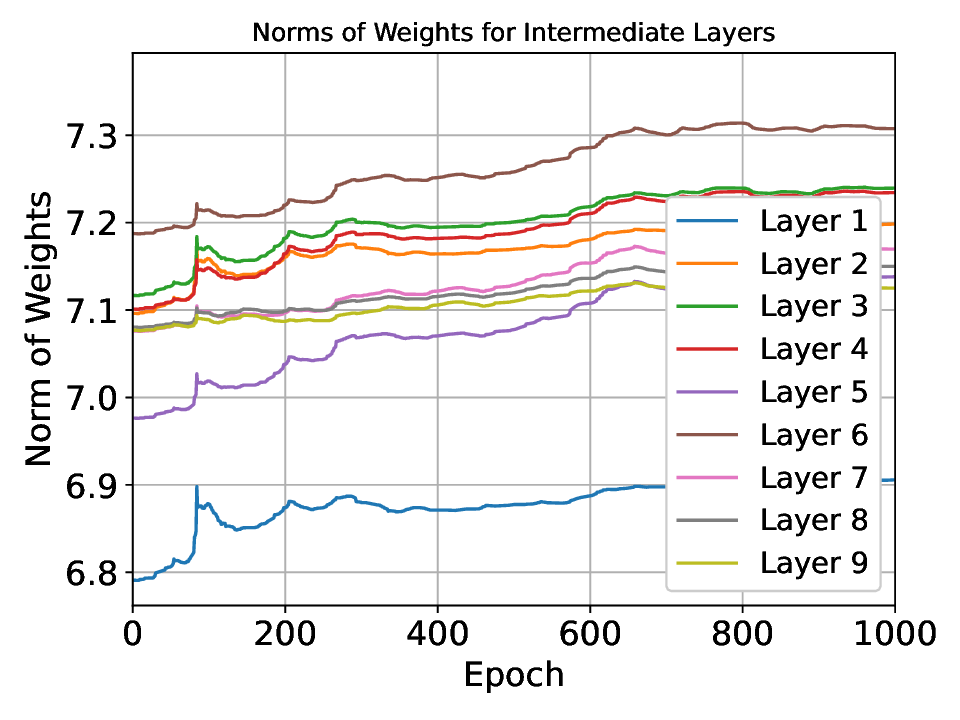}}
%\subfigure[Plain NN]{ \label{F1_n500_50x10_PCA_pnn}
%\includegraphics[width=2.95in]{F1_n500_50x10_PCA_pnn}}
\subfigure[SQR-SkipResNet]{ \label{F1_n500_50x10_normW_rsn}
\includegraphics[width=2.95in]{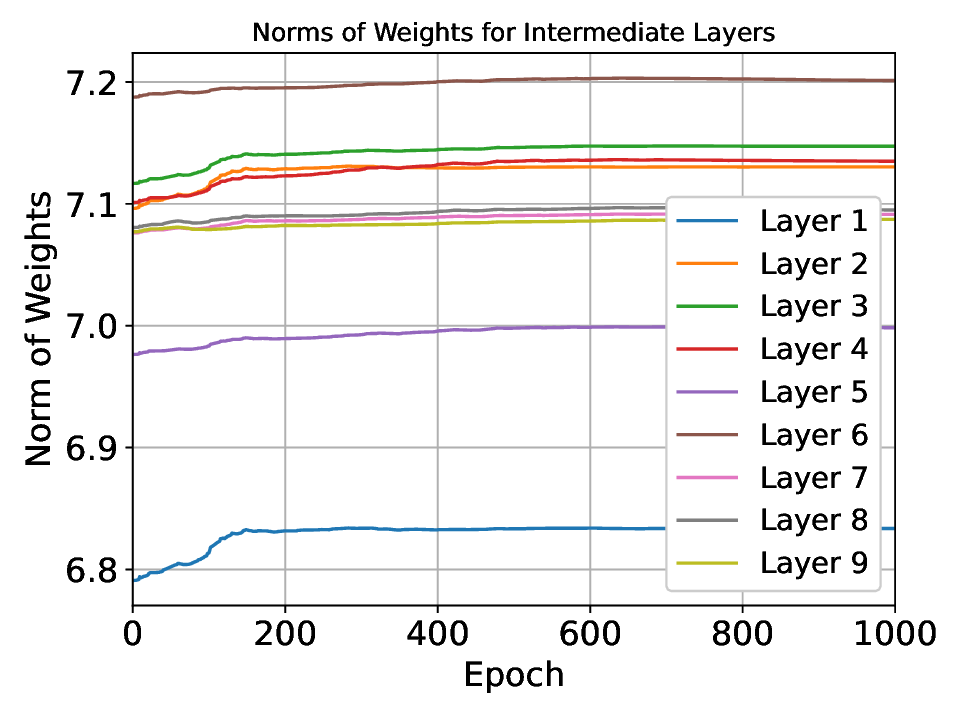}}
%\subfigure[SQR-SkipResNet]{ \label{F1_n500_50x10_PCA_rsn}
%\includegraphics[width=2.95in]{F1_n500_50x10_PCA_rsn}}
\caption{The profile of norm of weights both Plain NN (left panel) and SQR-SkipResNet (right panel) } \label{Ex1_5}
\end{figure}

\begin{figure}
\centering%
\subfigure[layer 1]{ \label{F1_n500_50x10_histL1_n}
\includegraphics[width=1.95in]{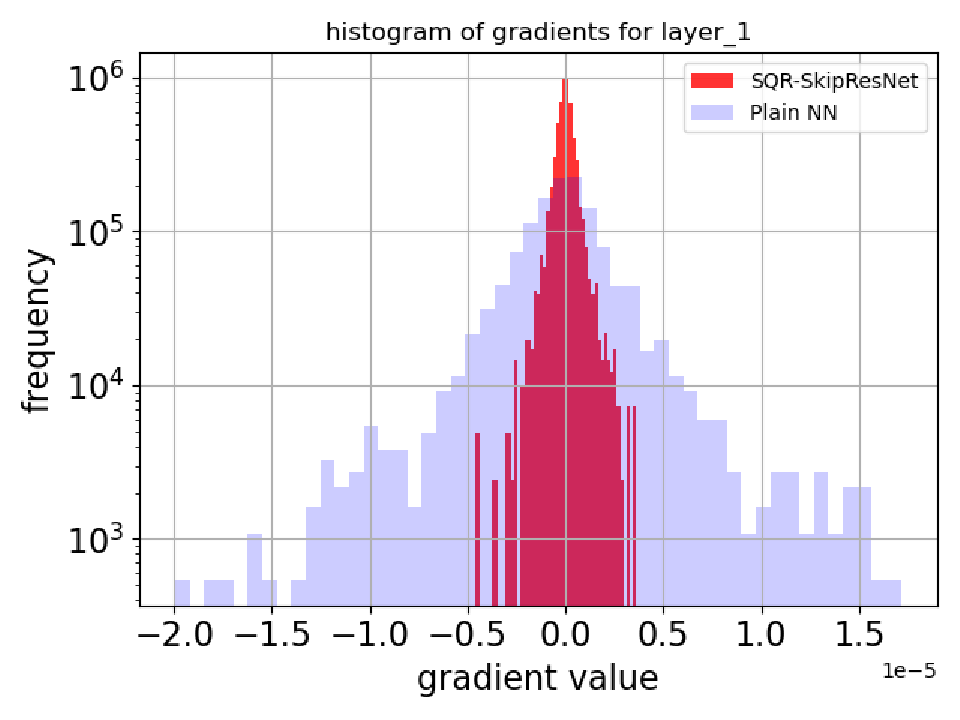}}
\subfigure[layer 5]{ \label{F1_n500_50x10_histL5_n}
\includegraphics[width=1.95in]{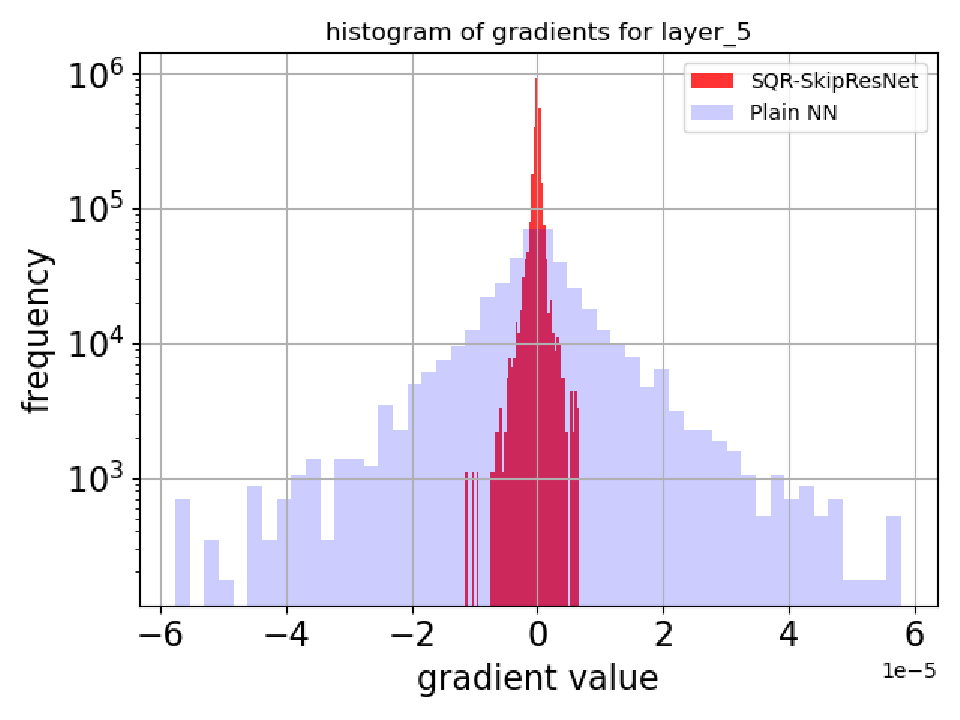}}
\subfigure[layer 9]{ \label{F1_n500_50x10_histL9_n}
\includegraphics[width=1.95in]{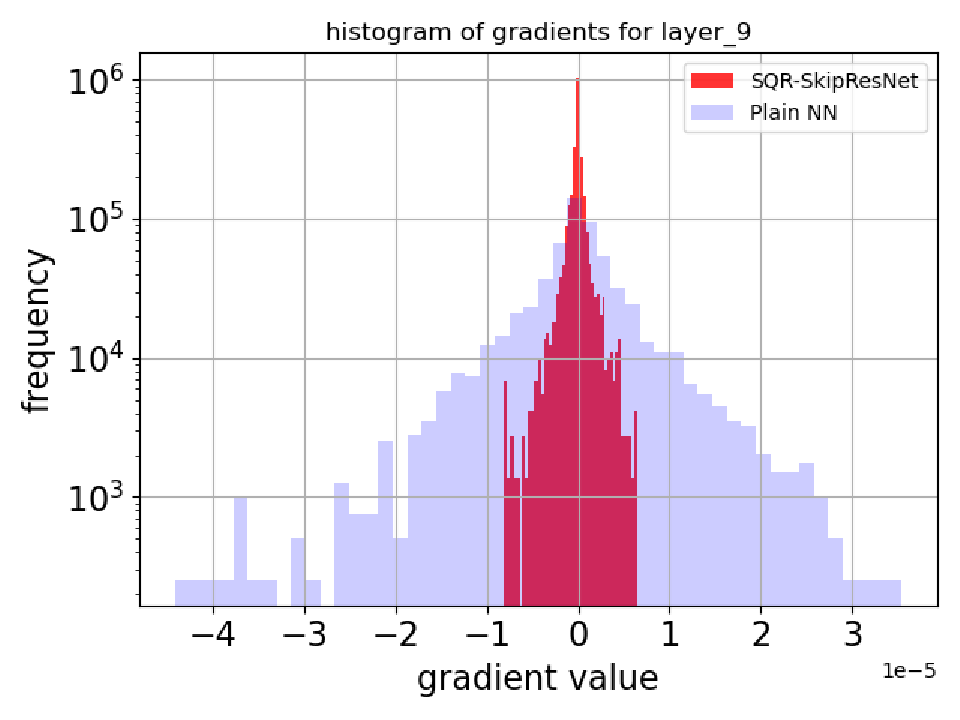}}
\caption{The histogram of gradient of loss with respect to the weight. } \label{Ex1_6}
\end{figure}

To elucidate the factors contributing to the differences between the Plain NN and SQR-SkipResNet, we conducted an analysis of the Frobenius norm of the weights updated throughout 1000 epochs across all hidden layers.
For this analysis, the Frobenius norm, represented as
\begin{equation}
 ||\textbf{W}||_F = \sqrt{\sum_{i=1}^{\mathcal{M}} \sum_{j=1}^{\mathcal{N}} |W_{ij}|^2}
\end{equation}
was chosen due to its suitability for capturing the overall magnitude and fluctuations of weight matrices. In this expression, $\mathcal{M}$ presents the number of epochs and $\mathcal{N}$ indicates the number of weight for all layers at one specific epoch. The resulting plots, depicted in Fig.~\ref{Ex1_5}, showcase distinct patterns in the evolution of weight norms between the two models. Specifically, Fig.~\ref{F1_n500_50x10_normW_pnn} illustrates the fluctuating behavior of weight norms in the Plain NN, whereas Fig.~\ref{F1_n500_50x10_normW_rsn} demonstrates a more stable trend in the SQR-SkipResNet model.

Notably, the SQR-SkipResNet model exhibits a convergence to stable weight norms after approximately 200 epochs, while the Plain NN continues to experience rising norms until the end of the 1000 epochs.
This divergence in weight behavior can be attributed to the fact that the residual connections in SQR-SkipResNet facilitate smoother optimization by providing clear paths for the flow of gradients \cite{He16}, allowing for easier weight updates. This smoother optimization process contributes to the observed stability in the evolution of weight norms in the SQR-SkipResNet model.
Therefore, the introduction of skip connections and residual connections in the SQR-SkipResNet architecture plays a crucial role in addressing optimization challenges encountered in deeper networks, leading to more stable weight norms and faster convergence compared to the Plain NN.

Furthermore, in order to investigate the underlying cause of the Plain NN's inability to deliver accurate predictions, we refer to the seminal works of Glorot and Bengio \cite{Glorot10} and Wang et al. \cite{Wang21}. We analyze the distribution of back-propagated gradients concerning the neural network weights \eqref{bp} throughout the training process, as depicted in Fig.\ref{Ex1_6} for various hidden layers ($n_l = 1, 5,$ and $9$) in Fig.\ref{F1_n500_50x10_histL1_n}, Fig.\ref{F1_n500_50x10_histL5_n}, and Fig.\ref{F1_n500_50x10_histL9_n}, respectively.
We observe that the back-propagated gradients for SQR-SkipResNet are smaller in magnitude compared to the Plain NN, indicating smoother gradient flow during training. This phenomenon is attributed to the presence of skip connections in ResNet, which enable gradients to bypass multiple layers, thereby facilitating more efficient optimization as initially introduced by He et al. \cite{He15, He16}. Additionally, the smaller back-propagated gradients in SQR-SkipResNet suggest better convergence compared to the Plain NN, attributed to a smoother loss surface as shown by Li et al. \cite{Li18}.
 Consequently, SQR-SkipResNet demonstrates enhanced training stability, faster convergence, and ultimately, improved accuracy across.

More investigations on the performance of the SQR-SkipResNet has been done by interpolating the non-smooth functions F2 and F3. Figure~\ref{Ex1_3} presents the interpolation results for F2 on the top panel and F3 on the bottom panel with $n=1000$. The corresponding training and validation error with respect to the epoch are shown in the first column. The second and third columns show the interpolated surface using Plain NN and SQR-SkipResNet, respectively.  Clearly, a better surface interpolation  has been carried out using the proposed method. More details are listed in Table~\ref{tab:ex1b}. This table shows that the accuracy using the SQR-SkipResNet is slightly better than Plain NN, however it is worth nothing that these functions are non-smooth and a slightly changes in error would affect the quality of interpolation tremendously as shown in Fig.~\ref{Ex1_3}. However to reach a better accuracy, the SQR-SkipResnet requires larger number of iterations and consequently the higher CPU time. This can be seen as the trade-off that SQR-SkipResNet makes for interpolating non-smooth functions to obtain better accuracy, in contrast to the smaller CPU time it requires for interpolating smooth functions.

\begin{figure}
\centering%
\subfigure[]{ \label{Ex2_F2_n1000_50x20}
\includegraphics[width=1.95in]{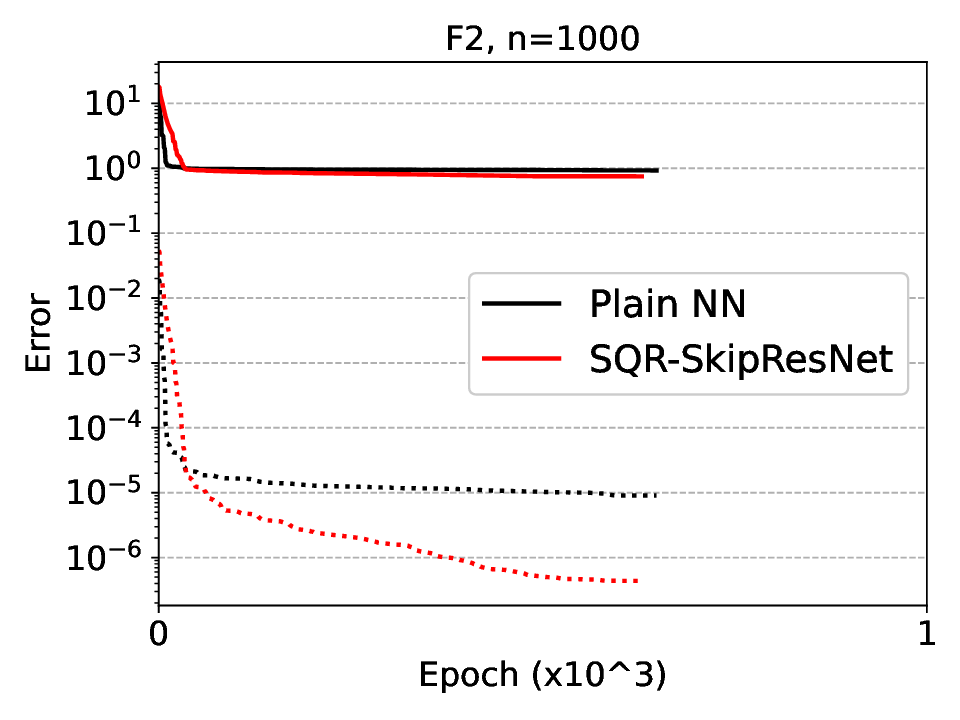}}
\subfigure[]{ \label{Ex2_F2_n1000_50x20_Surf_pnn}
\includegraphics[width=1.95in]{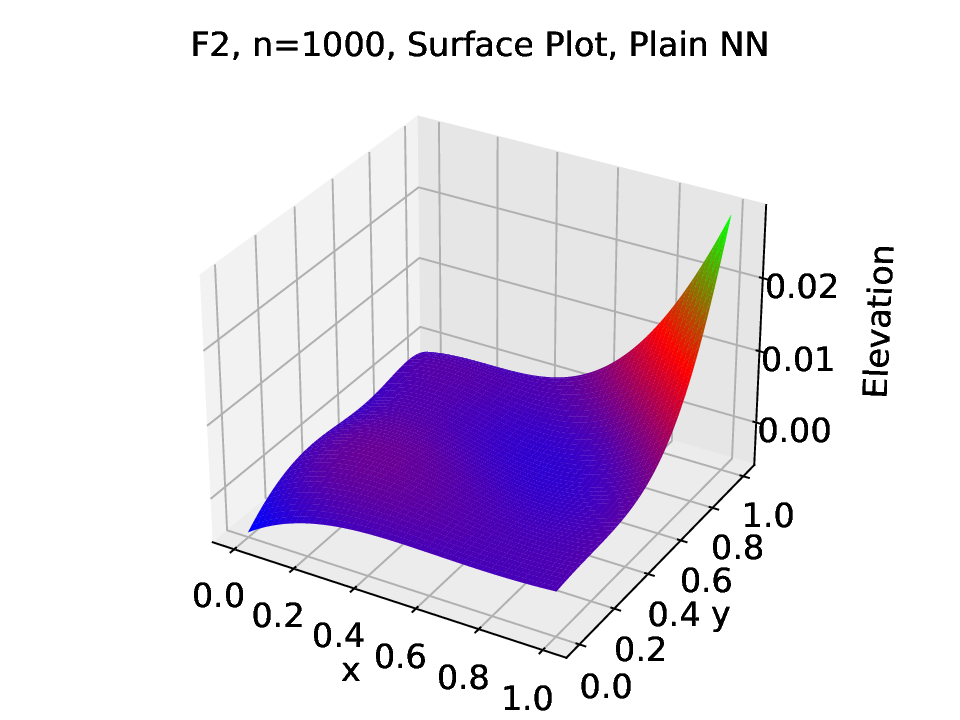}}
\subfigure[]{ \label{Ex2_F2_n1000_50x20_Surf_sqr}
\includegraphics[width=1.95in]{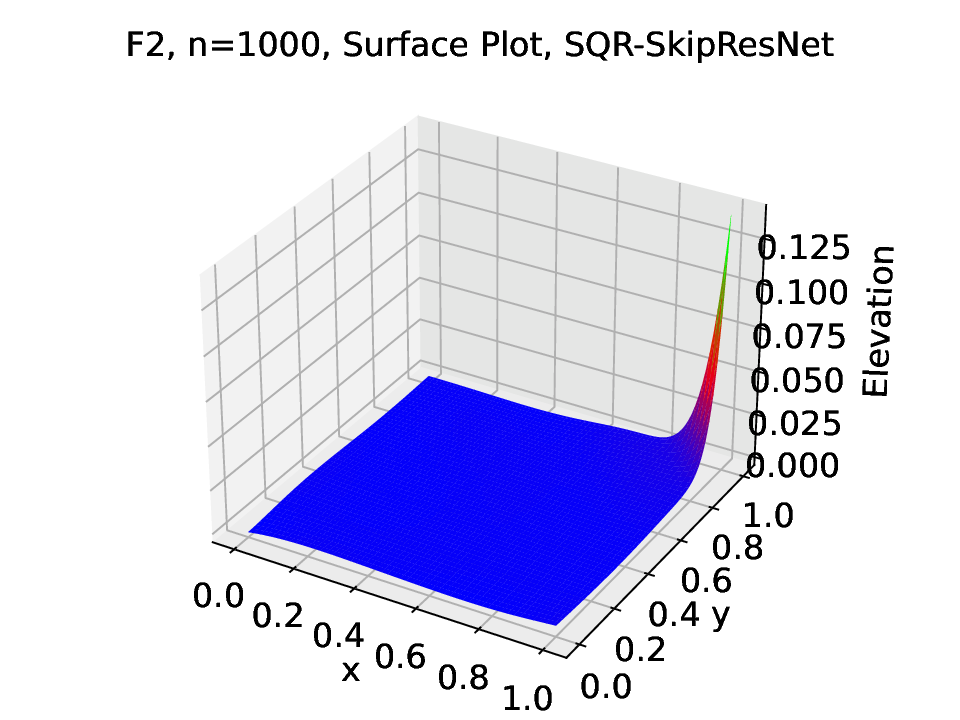}}
\subfigure[]{ \label{Ex2_F3_n1000_50x20}
\includegraphics[width=1.95in]{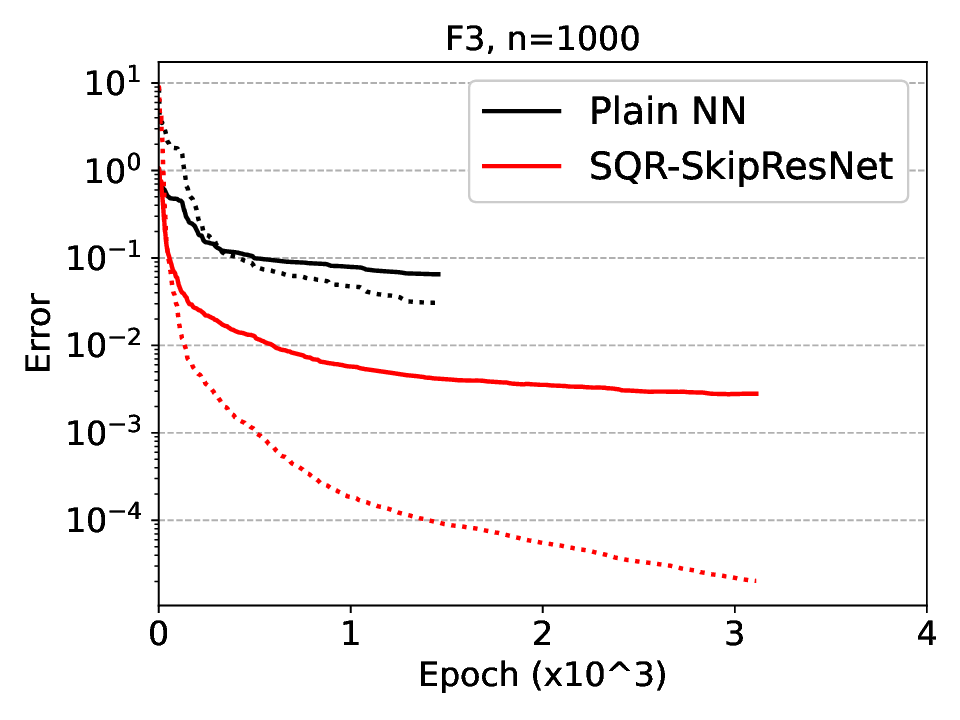}}
\subfigure[]{ \label{Ex2_F3_n1000_50x20_Surf_pnn}
\includegraphics[width=1.95in]{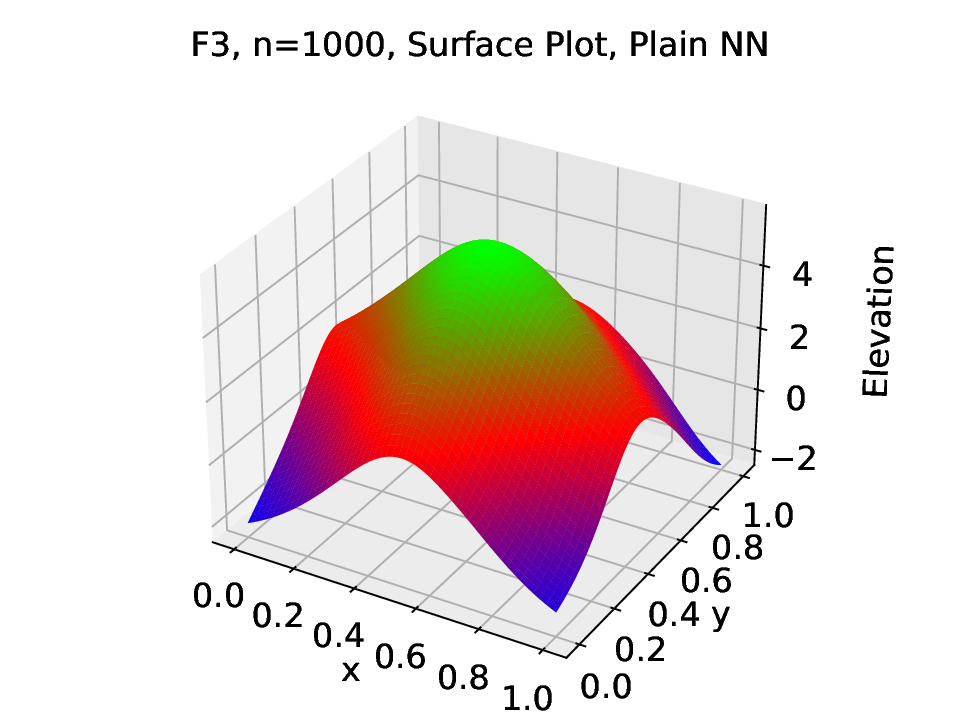}}
\subfigure[]{ \label{Ex2_F3_n1000_50x20_Surf_sqr}
\includegraphics[width=1.95in]{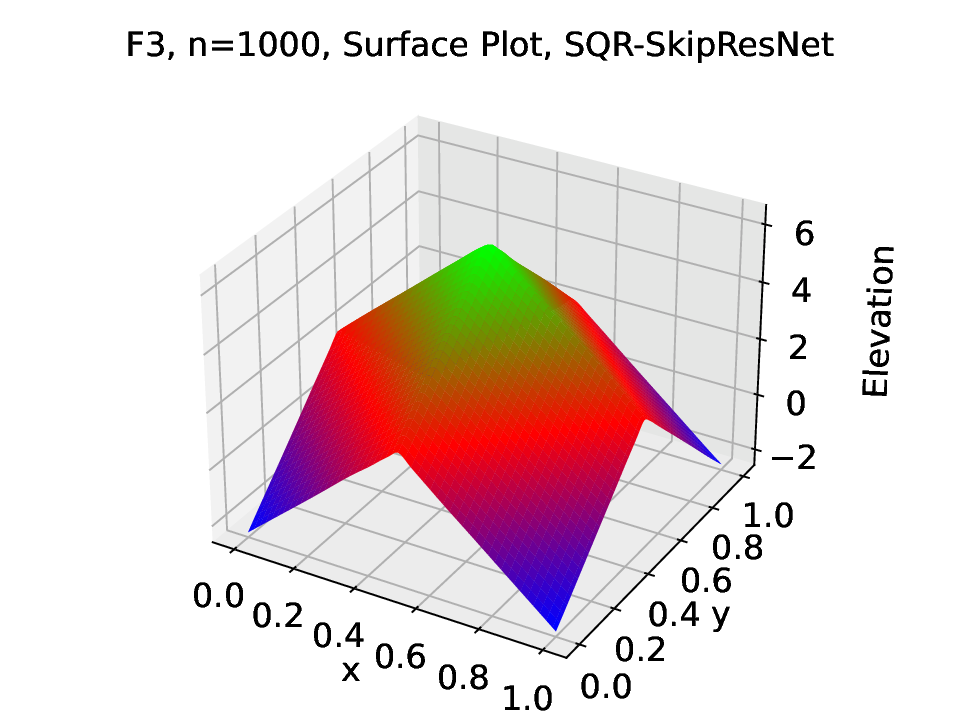}}
\caption{Example 1: The profile of the F2, and F3 } \label{Ex1_3}
\end{figure}

\begin{table}
\caption{Example 1: Maximum absolute errors and CPU time (t) for F2-F3.}\label{tab:ex1b}
\begin{center}
\begin{tabular}{p{1.5cm}cccc} \hline
 \multirow{2}{*}{function}& \multicolumn{2}{c}{\underline{\,\,\,\,\,\,\,\,\,\,\,\,\,\,Plain NN\,\,\,\,\,\,\,\,\,\,\,\,\,\,}}  & \multicolumn{2}{c}{\underline{\,\,\,\,\,\,\,SQR-SkipResNet\,\,\,\,\,\,\,}}
\\
 &error&t(s)        & error& t(s)
\\ \hline \hline
F2 &9.71e-01  &22     &8.59e-01  &27\\
F3 &7.98e-01 &51     &8.57e-02  &132\\
 \hline
%\end{tcolorbox}
\end{tabular}
\end{center}
\end{table}

}
\end{example}

%%%%%%%%%%%%%%%%%%%%%%%%%%%%%%%%%%%%%%%%%%%%%%%%%%%
%%%%%%%%%%%%%%%%%%%%%%%%%%%%%%%%%%%%%%%%%%%%%%%%%
%%%%%%%%%%%%%%%%%%%%%%%%%%%%%%%%%%%%%%%%%%%%%%%%%%%%
%%%%Example 4
\begin{example}\rm{
In this example, we demonstrate the performance of the proposed method in a real case study. Specifically, we interpolate the Mt. Eden or Maungawhau volcano in Auckland, NZ, as depicted in Fig.~\ref{Mt_Eden1} \cite{Eden}. The available data consists of 5307 elevation points uniformly distributed in a mesh grid area of size 10 by 10 meters \red{\cite{R14,Cavoretto21}}. Plot (b) shows the 3D surface, and plot (c) presents the contour plot of the volcano. Reference \cite{Cavoretto21} uses various radial basis functions (RBFs) with 118 collocation points to approximate the interpolated function for this example. The authors used leave-one-out cross-validation (LOOCV) approaches to determine the uncertainties in the RBF method. They show that a plain LOOCV, depending on the type of RBF, can lead to maximum absolute errors ranging from 12.6 to 54.7. With this background knowledge, we aim to solve this problem using deep learning.

\begin{figure}
\centering%
\subfigure[]{ \label{Mt_Eden1}
\includegraphics[width=1.8in]{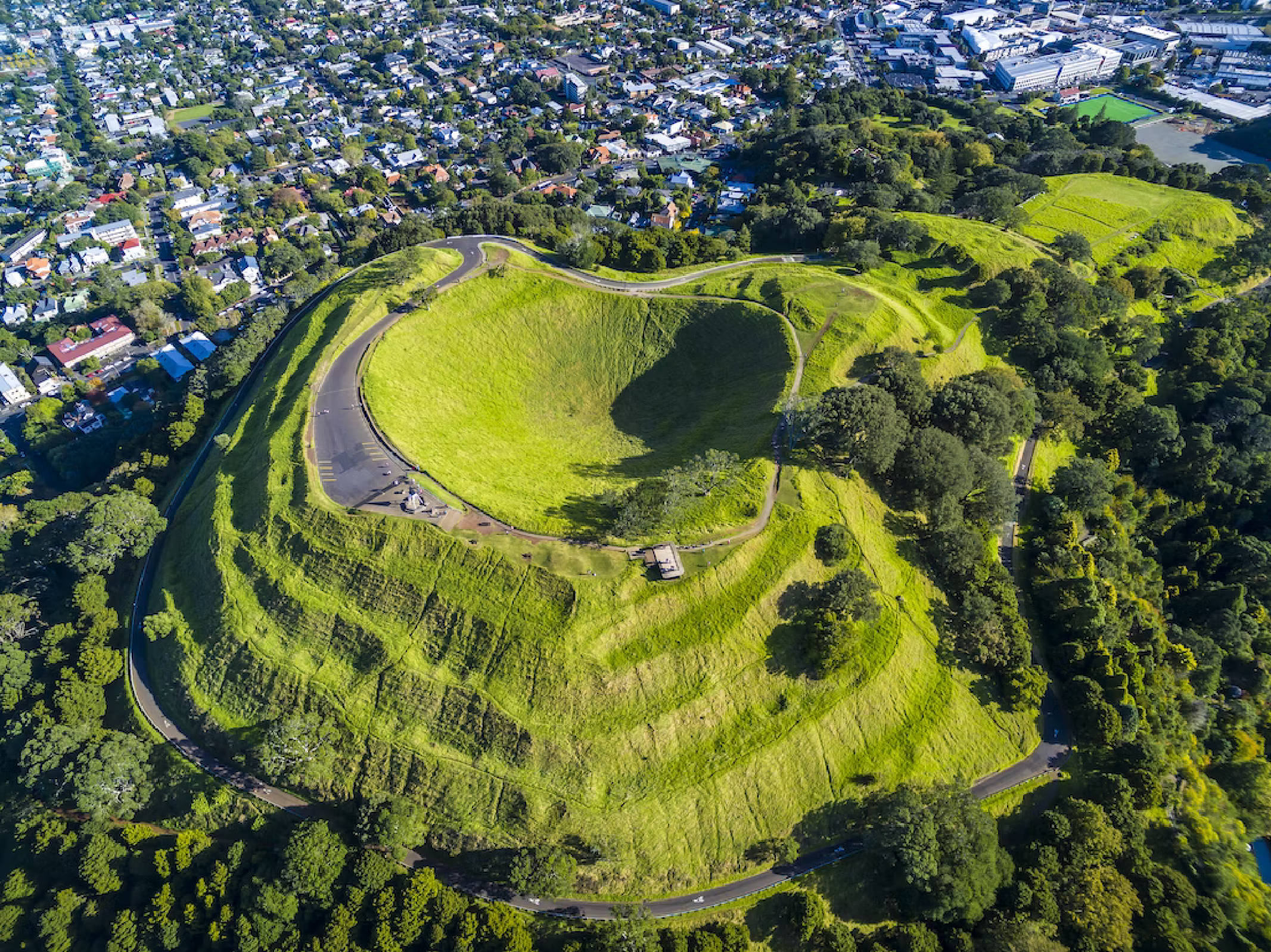}}
\subfigure[]{ \label{FMT_exact_3D}
\includegraphics[width=2.0in]{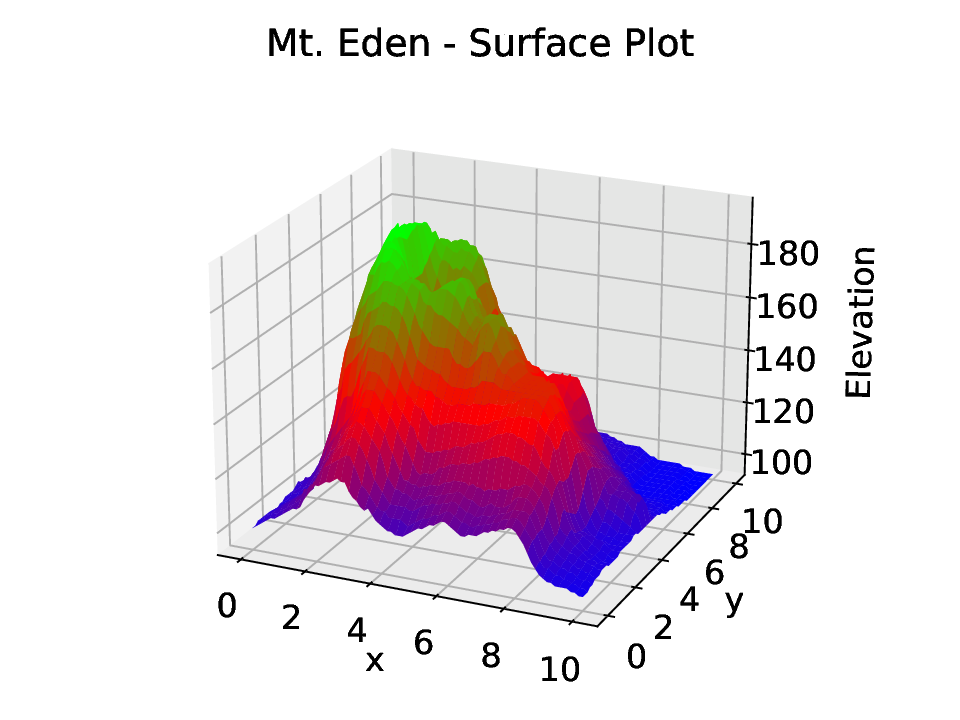}}
\subfigure[]{ \label{FMT_exact_2D}
\includegraphics[width=2.0in]{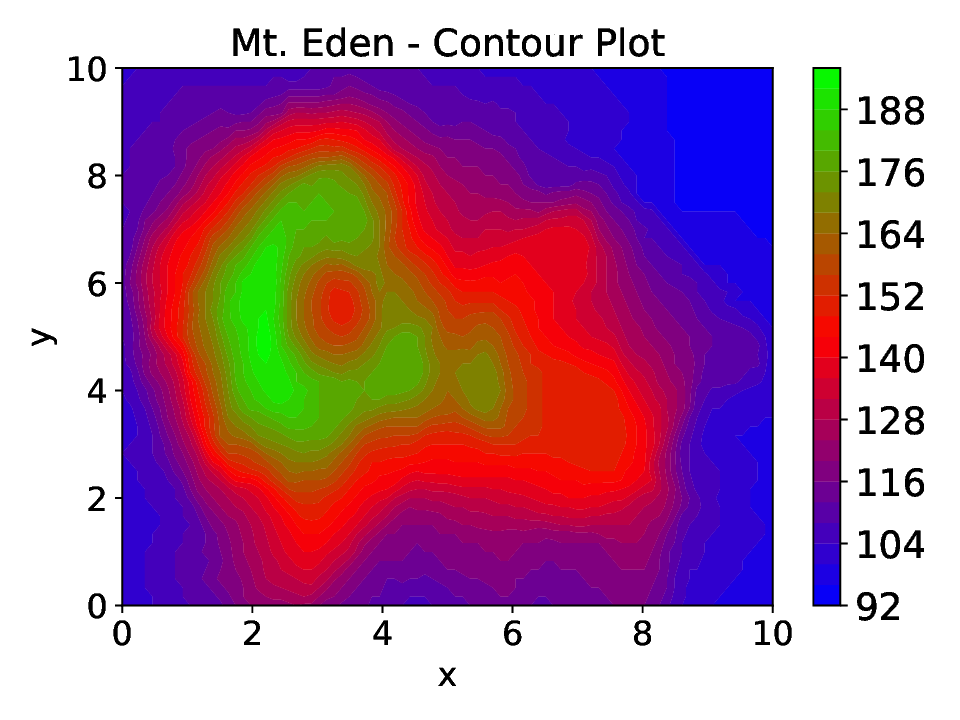}}
\caption{Example 2: (a) An image showcasing the Mt. Eden or Maungawhau volcano located in Auckland, New Zealand \cite{Eden}. (b) A 3D surface representation generated from a dataset containing $n=5307$ data points. (c) A contour plot providing insights into the topography of Mt. Eden.} \label{Ex6_0}
\end{figure}

\begin{figure}
\centering%
\subfigure[]{ \label{Ex6_n200_100x5_loss}
\includegraphics[width=2.2in]{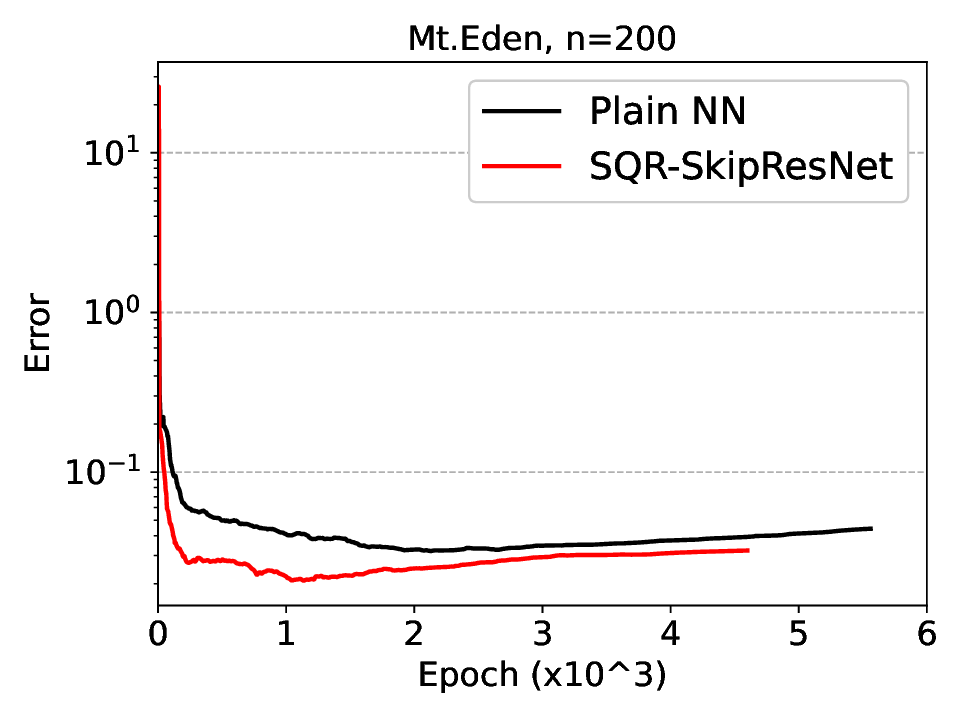}}\\
\subfigure[]{ \label{Ex6_n200_100x5_surf_pnn}
\includegraphics[width=1.9in]{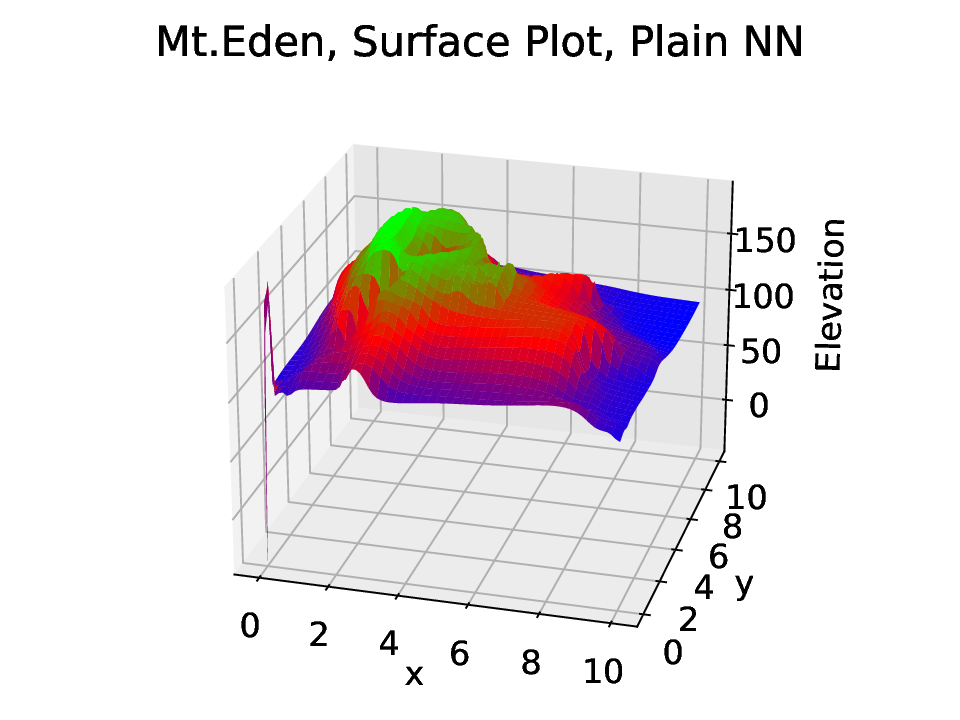}}
\subfigure[]{ \label{Ex6_n200_100x5_cont_pnn}
\includegraphics[width=1.9in]{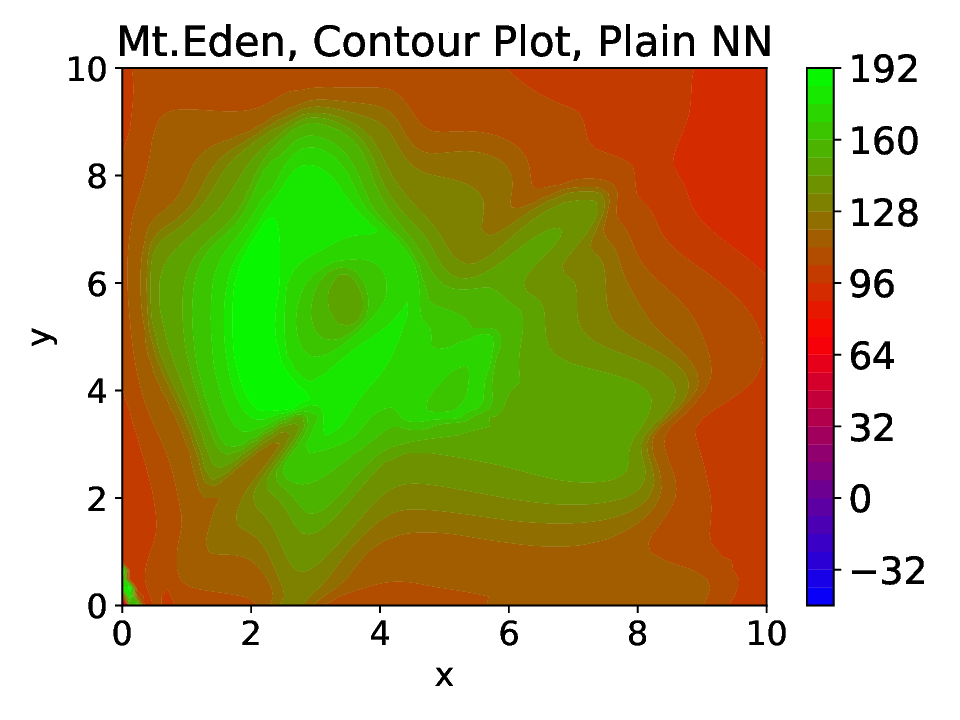}}
\subfigure[]{ \label{Ex6_n200_100x5_err_pnn}
\includegraphics[width=1.9in]{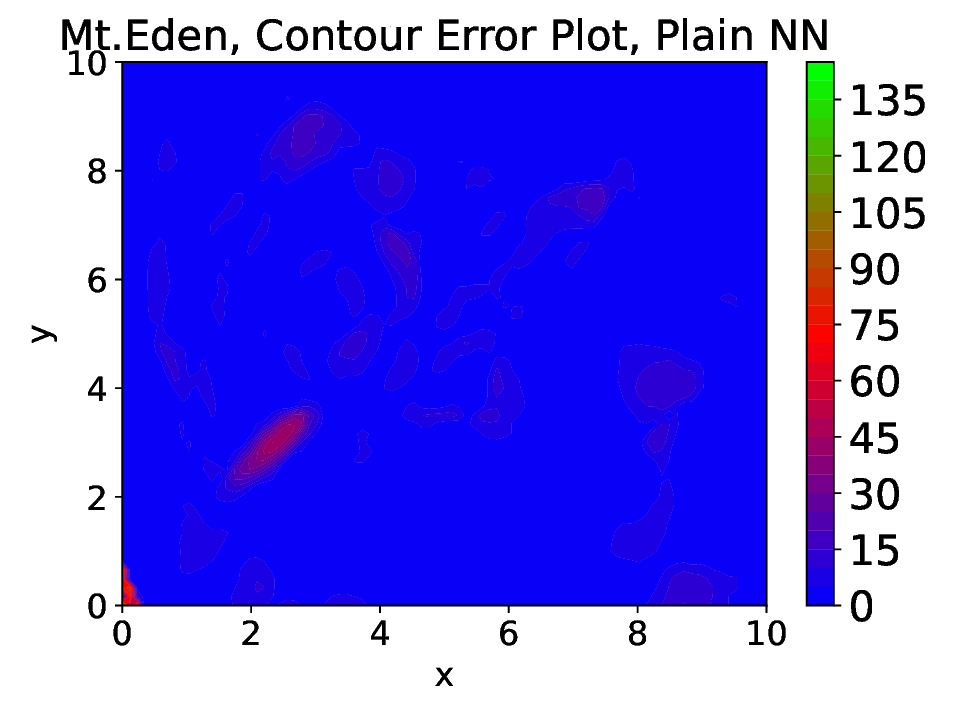}}
\subfigure[]{ \label{Ex6_n200_100x5_surf}
\includegraphics[width=1.9in]{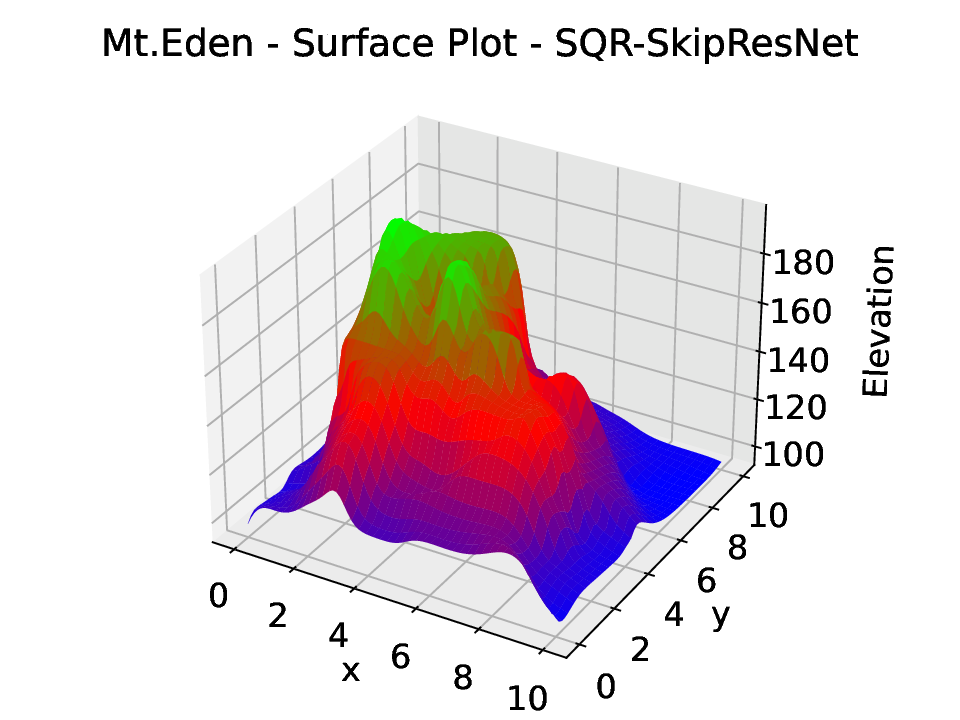}}
\subfigure[]{ \label{Ex6_n200_100x5_cont}
\includegraphics[width=1.9in]{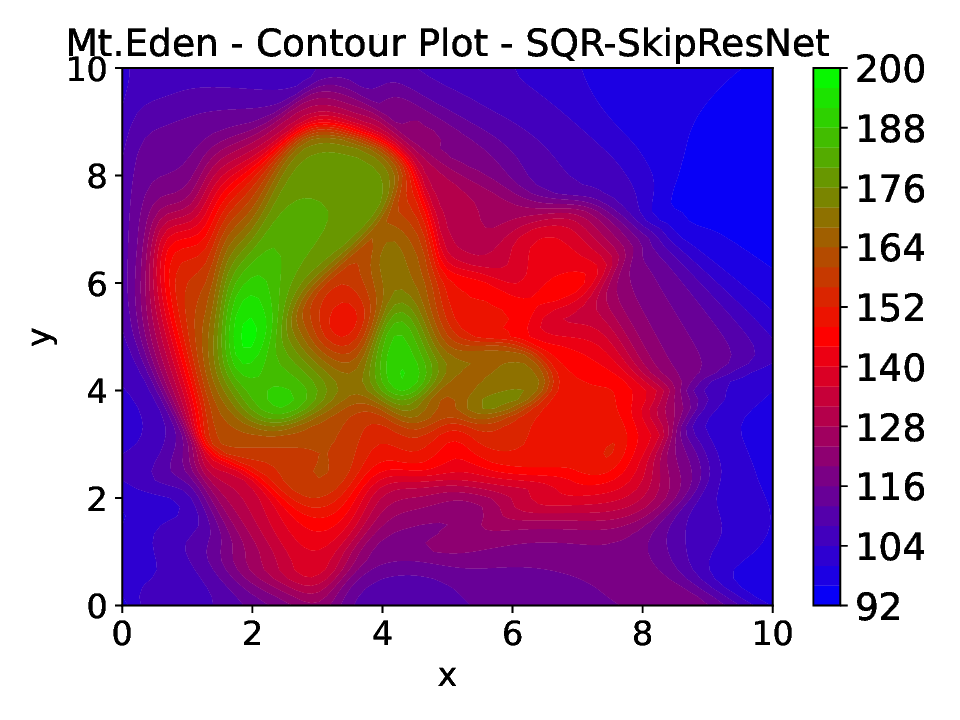}}
\subfigure[]{ \label{Ex6_n200_100x5_err}
\includegraphics[width=1.9in]{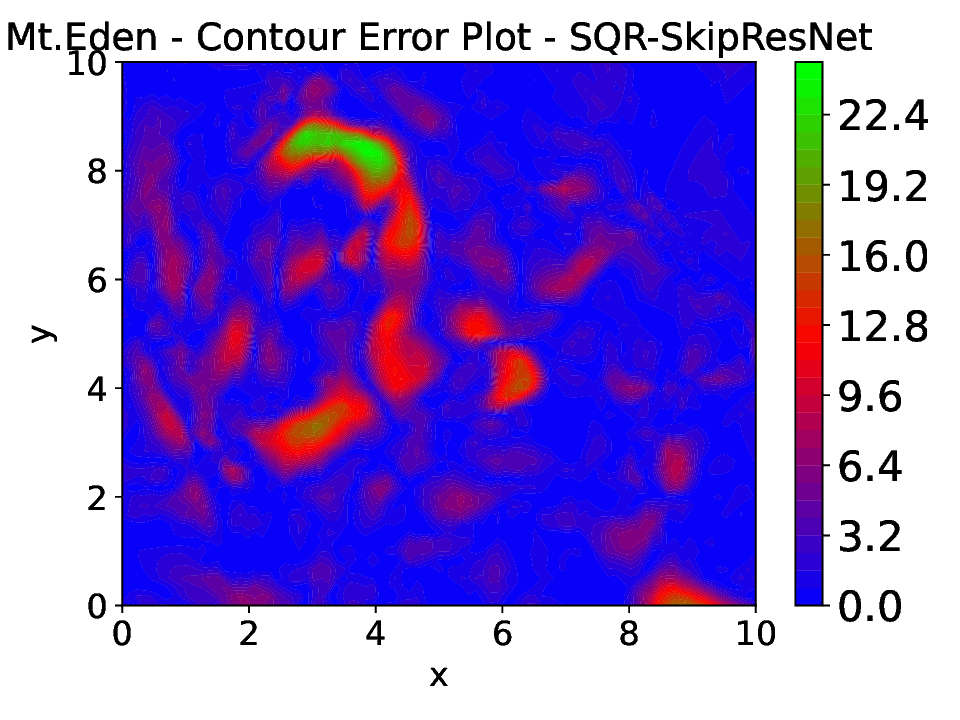}}
\caption{Example 2: Maximum absolute error for Mt. Eden interpolation using L-BFGS-B optimizer with $n=200$, $n_n=50$, and $n_l=5$.} \label{Ex6_1}
\end{figure}

\begin{figure}
\centering%
\subfigure[]{ \label{Ex6_n200_100x5_loss_adam}
\includegraphics[width=2.2in]{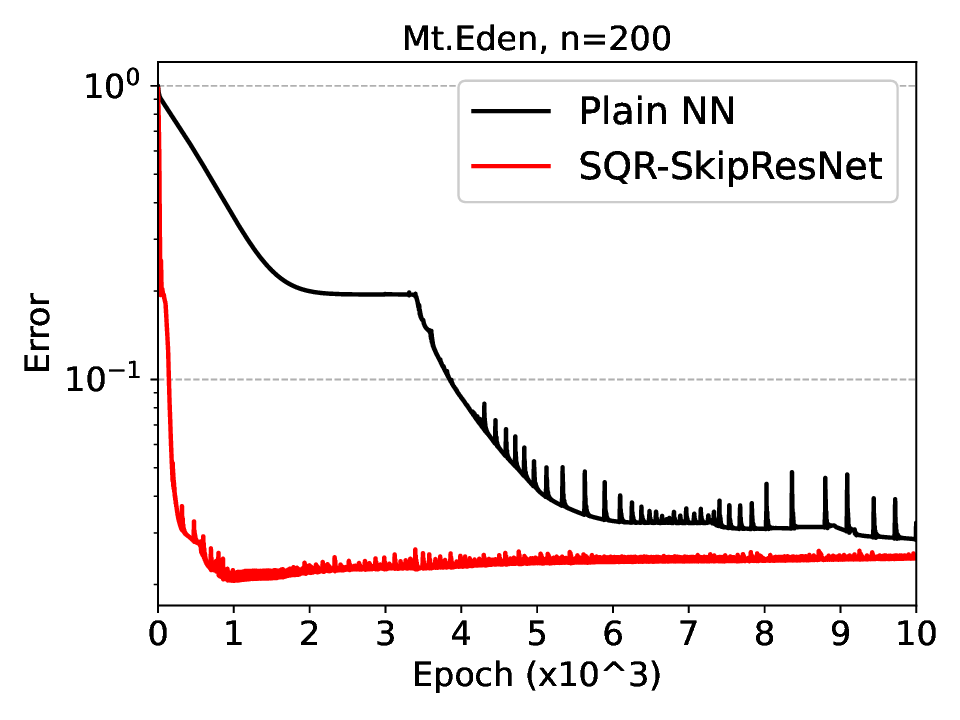}}\\
\subfigure[]{ \label{Ex6_n200_100x5_surf_pnn_adam}
\includegraphics[width=1.9in]{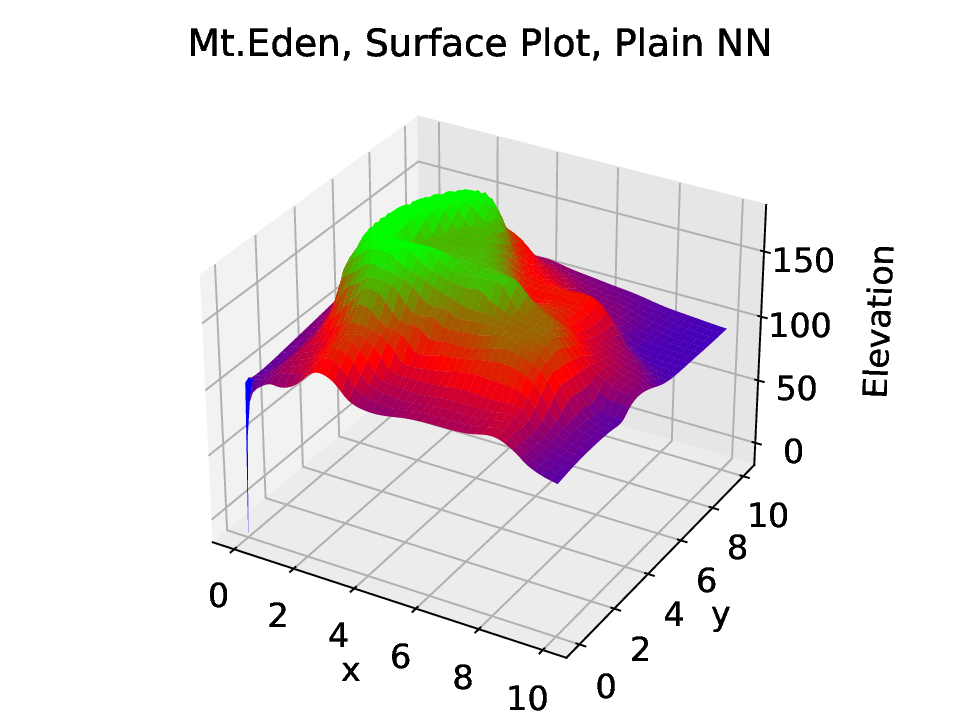}}
\subfigure[]{ \label{Ex6_n200_100x5_cont_pnn_adam}
\includegraphics[width=1.9in]{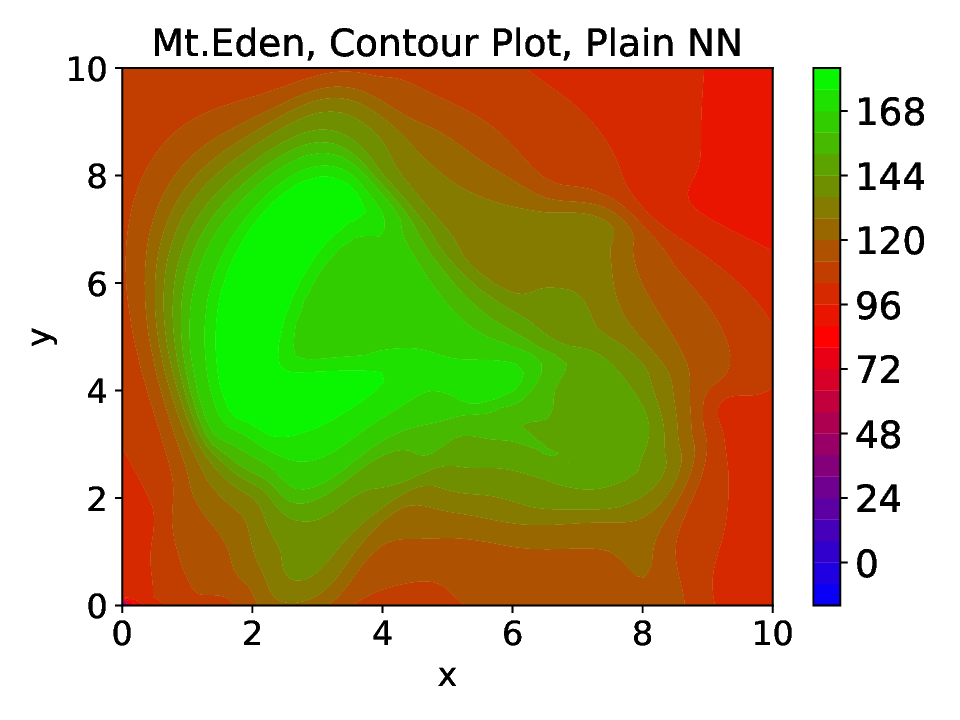}}
\subfigure[]{ \label{Ex6_n200_100x5_err_pnn_adam}
\includegraphics[width=1.9in]{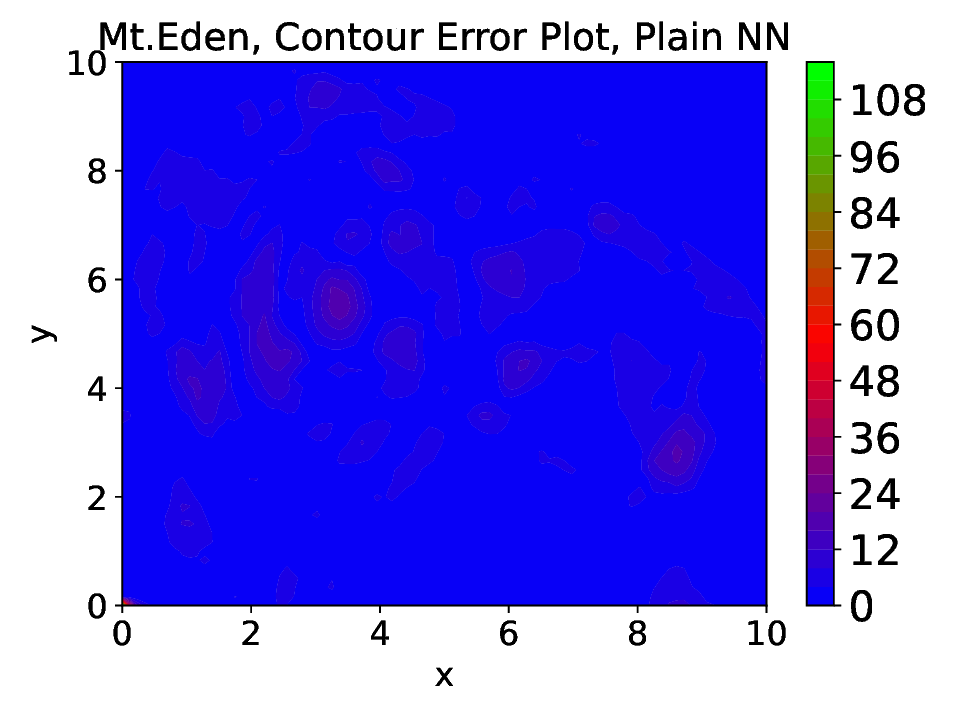}}
\subfigure[]{ \label{Ex6_n200_100x5_surf_adam}
\includegraphics[width=1.9in]{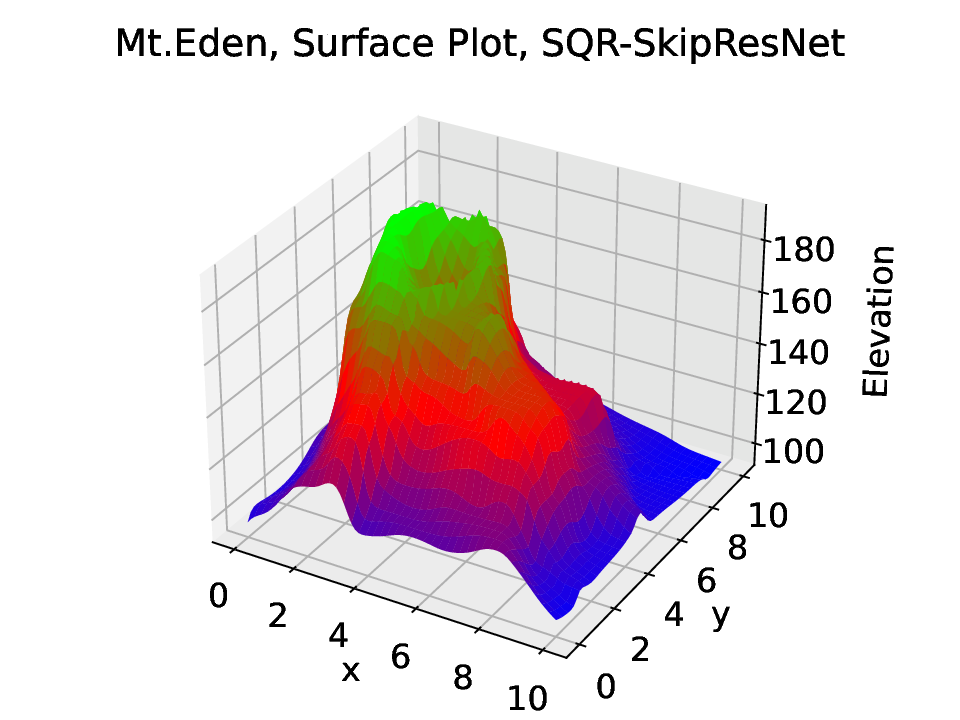}}
\subfigure[]{ \label{Ex6_n200_100x5_cont_adam}
\includegraphics[width=1.9in]{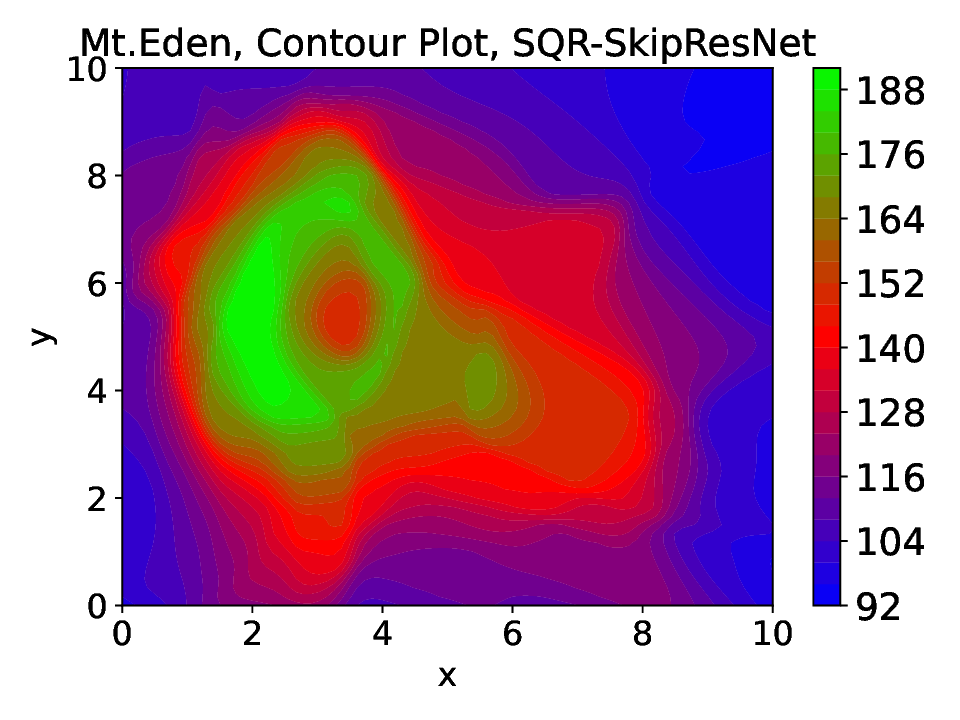}}
\subfigure[]{ \label{Ex6_n200_100x5_err_adam}
\includegraphics[width=1.9in]{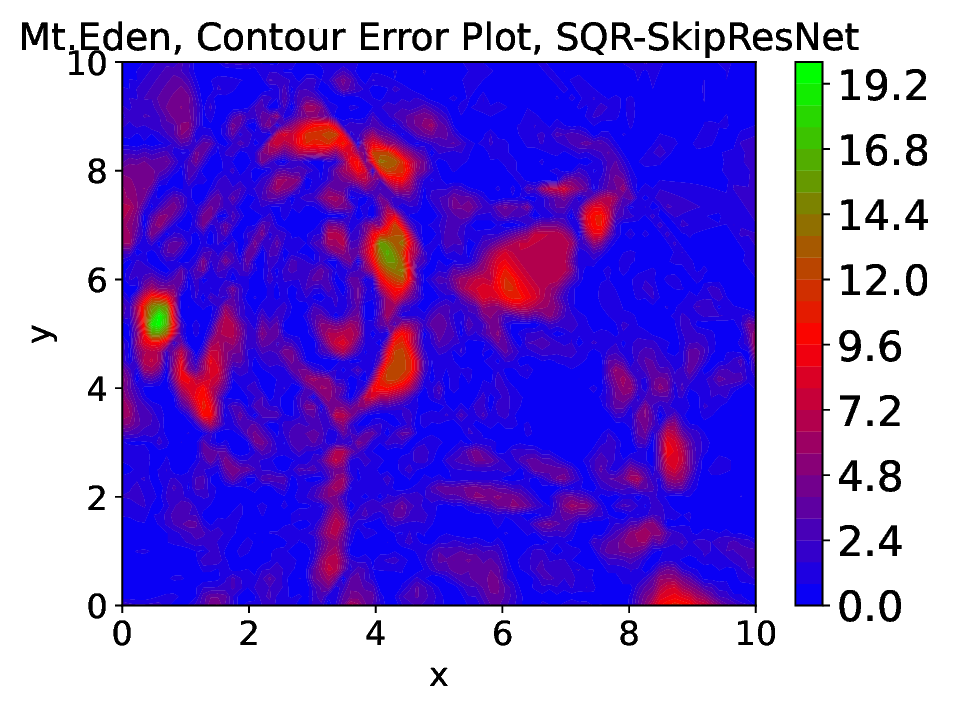}}
\caption{Example 2: Mt. Eden interpolation results using Adam optimizer with $n=200$, $n_n=50$, and $n_l=5$.} \label{Ex6_2}
\end{figure}

\noindent
In the first experiment, we utilize only 200 collocation points, 5 hidden layers with $n_n= 100$, and we optimize the training using L-BFGS-B. The remaining data, 5107 data points, are used for validation. The results are shown in Fig.~\ref{Ex6_n200_100x5_loss}, which illustrates the relative L2 norm error over the test data.
Evidently, SQR-SkipResNet achieves higher accuracy with fewer iterations. The convergence time for SQR-SkipResNet is 68 seconds, and it requires 4600 iterations to converge. On the other hand, Plain NN requires 80 seconds and 5600 iterations to achieve convergence.

Additionally, we provide more details on the interpolated surface and accuracy in Fig.~\ref{Ex6_1}. The second row shows the interpolated surface using Plain NN, while the third row shows the results obtained with SQR-SkipResNet. Specifically, plots \ref{Ex6_1}(b) and \ref{Ex6_1}(e) depict the interpolated surfaces for Plain NN and SQR-SkipResNet, respectively. Similarly, plots \ref{Ex6_1}(c) and \ref{Ex6_1}(f) display the contour plots for both methods. Finally, plots \ref{Ex6_1}(d) and \ref{Ex6_1}(g) represent the contour error plots, measured by the maximum absolute error, for Plain NN and SQR-SkipResNet, respectively.

Clearly, the results using SQR-SkipResNet significantly outperform those from Plain NN. The accuracy of Plain NN, specifically in terms of the maximum absolute error, improves significantly (500\%) when using the SQR-SkipResNet architecture. This underscores the superiority of SQR-SkipResNet in achieving more accurate and reliable interpolation results.

\begin{table}
\caption{Example 2:  Maximum absolute errors (\textit{m}) for Mt. Eden interpolation using Adam optimizer for various number of training data points $n$, neurons $n_n$ and layers $n_l$.}\label{tab:ex3a}
\begin{center}
\begin{tabular}{c|c|c|cc} \hline
$n$&$n_n$&$n_l$ & Plain NN & SQR-SkipResNet
\\
\hline \hline
 \multirow{4}{*}{200}
& \multirow{2}{*}{50}
&5 &\ding{55}      &12.9 \\
&&10 &\ding{55}
      &32.6 \\ \cline{2-5}
&\multirow{2}{*}{100}
&5  &114    &19.6  \\
&&10  &\ding{55}
    &25.5  \\ \hline
 \multirow{4}{*}{1000}
& \multirow{2}{*}{50}
&5 &21.6      &4.77 \\
&&10 &\ding{55}
      &14.0 \\  \cline{2-5}
&\multirow{2}{*}{100}
&5  &7.36    &6.28  \\
&&10  &\ding{55}
    &7.78  \\
 \hline
%\end{tcolorbox}
\end{tabular}
\end{center}
\end{table}

\begin{figure}
\centering%
\subfigure[Plain NN]{ \label{vpl_n200_50x10_normW_pnn}
\includegraphics[width=2.95in]{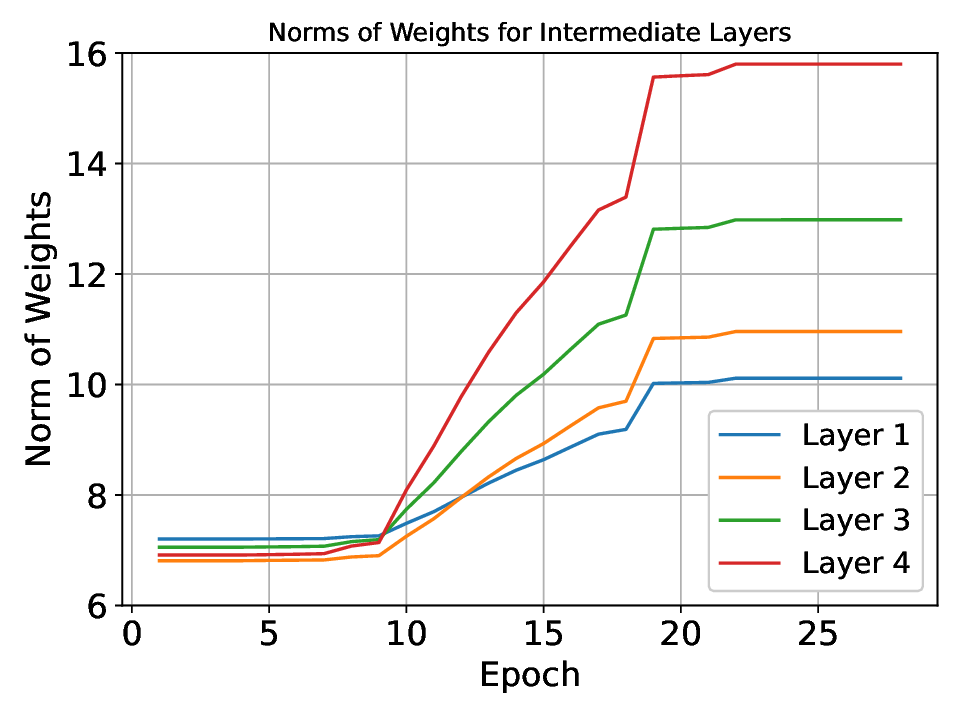}}
%\subfigure[Plain NN]{ \label{vol_n200_50x10_PCA_pnn}
%\includegraphics[width=2.95in]{vol_n200_50x10_PCA_pnn}}
\subfigure[SQR-SkipResNet]{ \label{vpl_n200_50x10_normW_rsn}
\includegraphics[width=2.95in]{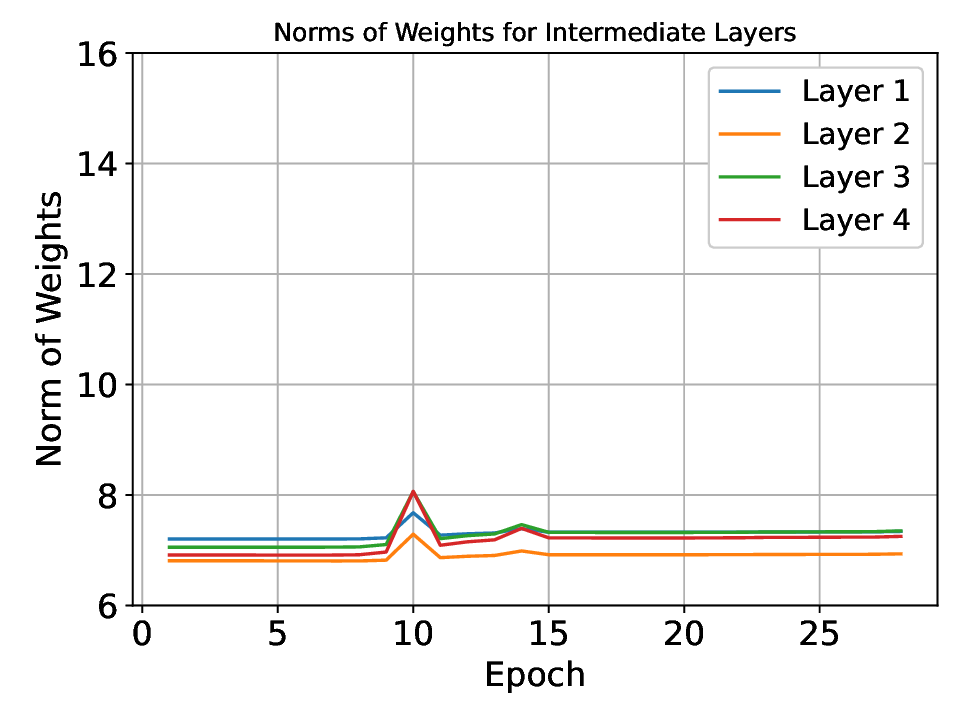}}
%\subfigure[SQR-SkipResNet]{ \label{vol_n200_50x10_PCA_rsn}
%\includegraphics[width=2.95in]{vol_n200_50x10_PCA_rsn}}
\caption{The profile of norm of weights both Plain NN (left panel) and SQR-SkipResNet (right panel). } \label{Ex2_5}
\end{figure}

In our second experiment, we repeat the the previous example but this time we use Adam optimizer with the learning rate of 1.0E-3, and 10k iteration. The organization of plots are as the previous example. Plot \ref{Ex6_2}(a) shows that SQR-SkipResNet works much more accurate from the beginning of the iterations with much less fluctuation compare with Plain NN (compare plots \ref{Ex6_2} (b) and \ref{Ex6_2}(e), respectively, and its corresponding contour plots in \ref{Ex6_2}(c) and \ref{Ex6_2}(f)). We also see that the interpolated surface when using the SQR-SkipResNet (plot \ref{Ex6_2}(g)) can be completely better than Plain NN (plot \ref{Ex6_2}(d)). The accuracy with respect to the maximum absolute error for the latter one is about 462\% better than the Plain NN. A comparisopn between these two optimizers, L-BFGS-B (Fig.~\ref{Ex6_1} and Fig.~\ref{Ex6_2}) shows a better performance using Adam for both Plain NN and proposed SQR-SkipResNet.

Therefore, we further investigate the impact of the number of data points $n$, neurons $n_n$, and layers $n_l$ as listed in Table~\ref{tab:ex3a}. In this table, \ding{55} denotes cases where training failed. When training fails, the interpolated surface remains partly flat and partly non-smooth.
 we have the following observations:

\begin{itemize}
\item As $n$ increases, smaller errors obtained.
\item With a fixed number of neurons $n_n$, the errors are smaller when the number of layers is $n_l=5$ compared to $n_l=10$.
\item With a fixed number of layers $n_l$, the errors are smaller when the number of neurons is $n_n=50$ compared to $n_n=100$.
\item Plain NN failed to train in 5 cases, while the proposed method exhibited successful performance.
\end{itemize}
Finally we see that in all cases, SQR-SkipResNet led to better accuracy compare to Plain NN.

Fig. \ref{Ex2_5} presents the Frobenius norm of the weights with respect to the epoch number for the first case in Table \ref{tab:ex3a} with $n=200$, $n_n=50$, and $n_l=5$. It is noteworthy that only 28 epochs are considered, as the Plain NN diverged thereafter. This training failure can be traced to the weight updates, where Fig. \ref{vpl_n200_50x10_normW_pnn} depicts a significant increase in weights after a few epochs. Conversely, the proposed Skip-SqrResNet algorithm, as shown in Fig. \ref{vpl_n200_50x10_normW_rsn}, exhibits more stable weight updating with respect to the epoch compared to the Plain NN.\\
The substantial changes in the norm of weights with respect to the epoch lead to the failure of the Plain NN. As observed in the figures, the norms of the weights for all layers typically range between 6 to 8, but at Epoch 28, they spike to values between 10 to 16. Such drastic changes can destabilize the learning process, causing the network to diverge. The sudden increase in weight norms indicates instability and erratic behavior in the learning dynamics, hindering the network's ability to converge to a satisfactory solution. Therefore, these large fluctuations in weight norms are detrimental to the training process and contribute to the failure of the Plain NN.

}
\end{example}

%%%%Example 4
\begin{example}\rm{
In the concluding example regarding the interpolation problems, we analyze the effectiveness of the proposed neural network in a 3D example, specifically using the Stanford bunny model \cite{Stan}, as depicted in Fig.~\ref{Ex7_bunnyFace}. The entire bunny model has been scaled by a factor of 10. A distribution of points over the bunny's surface is illustrated in Fig.~\ref{Ex7_bunnyPoints}, comprising a total of 8171 data points. The validation error is performed using the following test function (refer to \cite{Bozzini02}, F4):
\begin{align*}
{\rm F4}(x_1,x_2,x_3) = \frac{1}{3} \exp\left[-\frac{81}{16} \left((x_1-0.5)^2 + (x_2-0.5)^2 + (x_3-0.5)^2\right)\right].
\end{align*}

\begin{figure}
\centering%
\subfigure[]{ \label{Ex7_bunnyFace}
\includegraphics[width=2.55in]{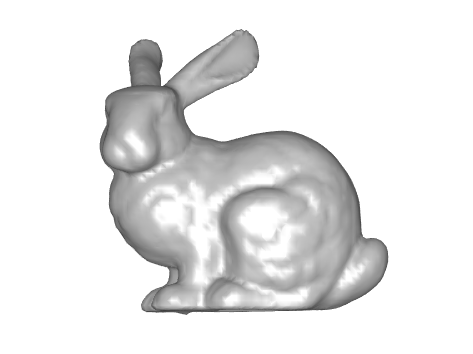}}
\subfigure[]{ \label{Ex7_bunnyPoints}
\includegraphics[width=2.55in]{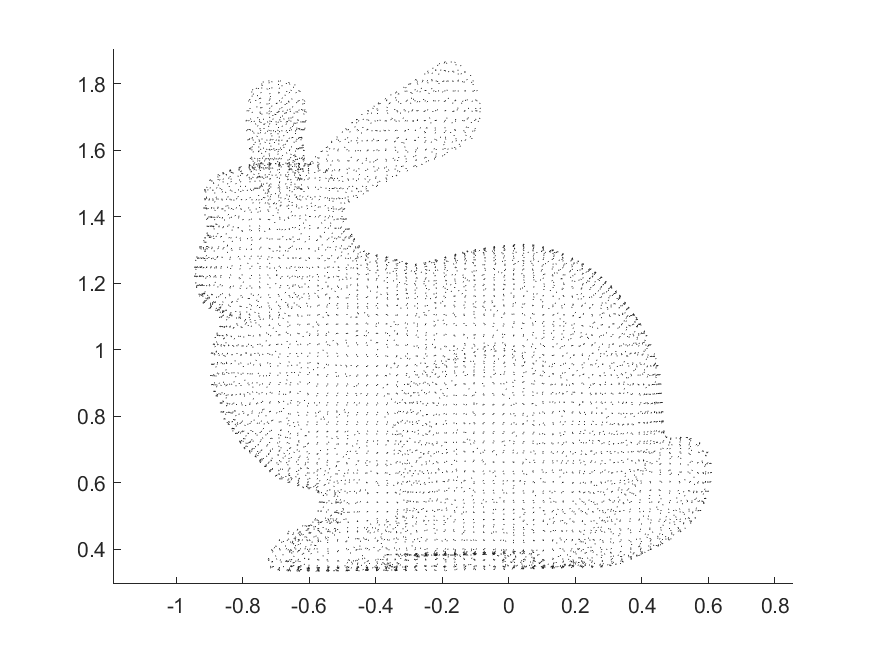}}
\caption{Example 3:  The Stanford Bunny {\cite{Stan}}. } \label{Ex7_1}
\end{figure}

 \begin{figure}
\centering%
\subfigure[]{ \label{Ex7_F4_n500_50x20}
\includegraphics[width=1.95in]{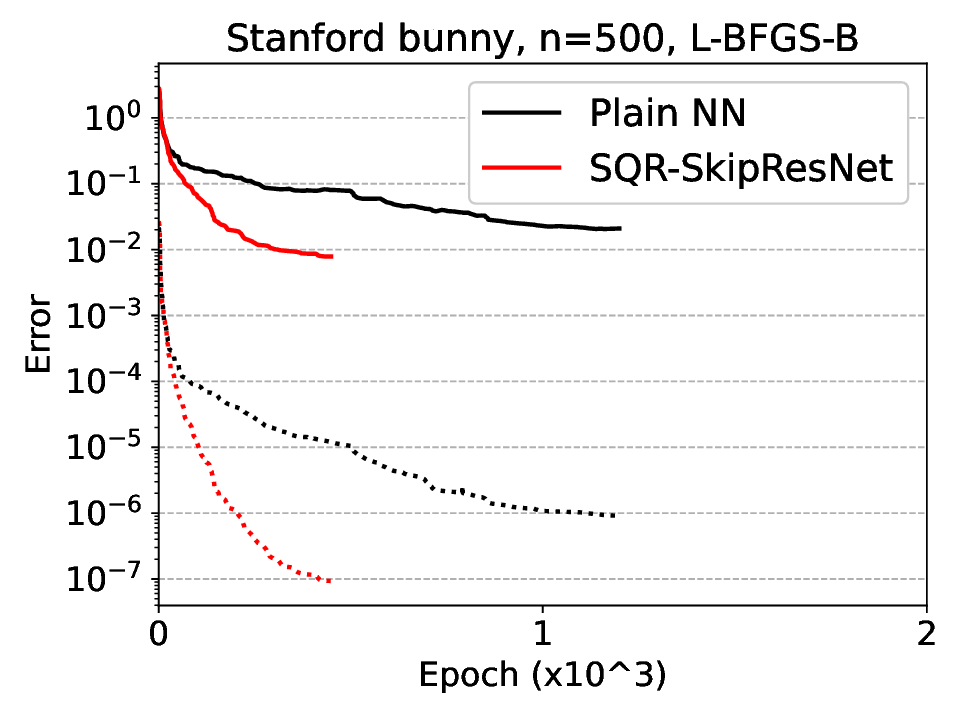}}
\subfigure[]{ \label{Ex7_F4_n500_50x20_pnn}
\includegraphics[width=1.95in]{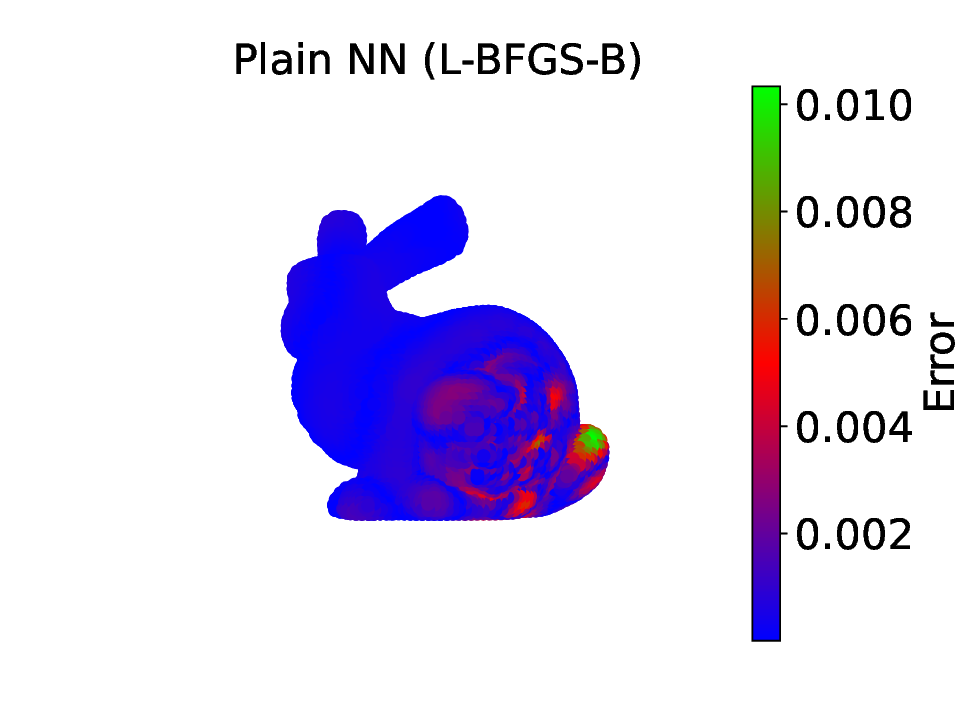}}
\subfigure[]{ \label{Ex7_F4_n500_50x20_sqr}
\includegraphics[width=1.95in]{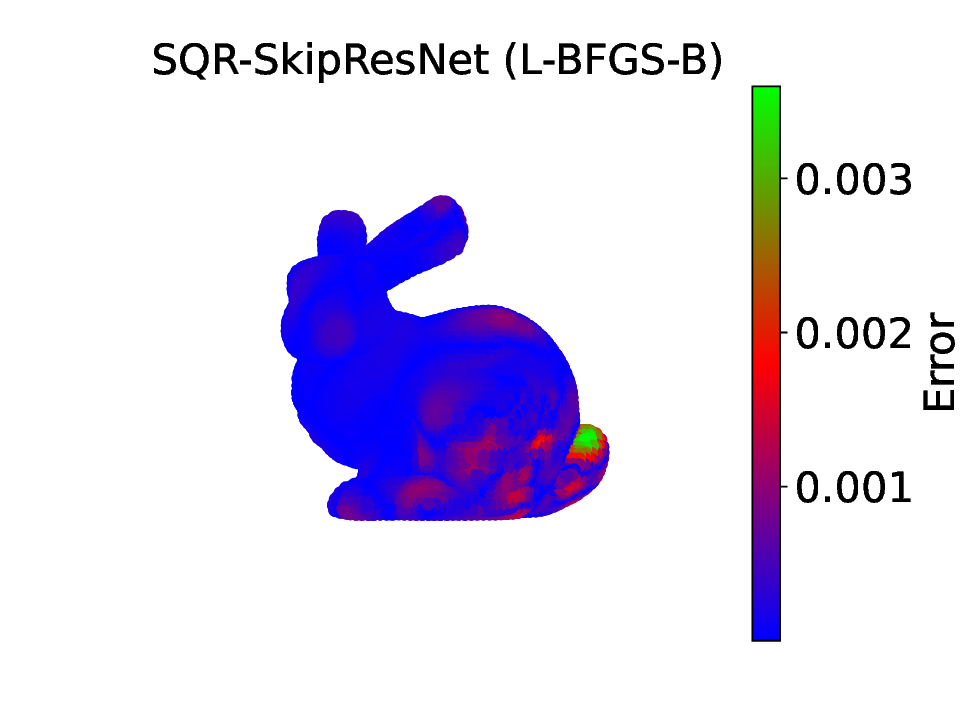}}
\subfigure[]{ \label{Ex7_F4_n500_50x20_adam}
\includegraphics[width=1.95in]{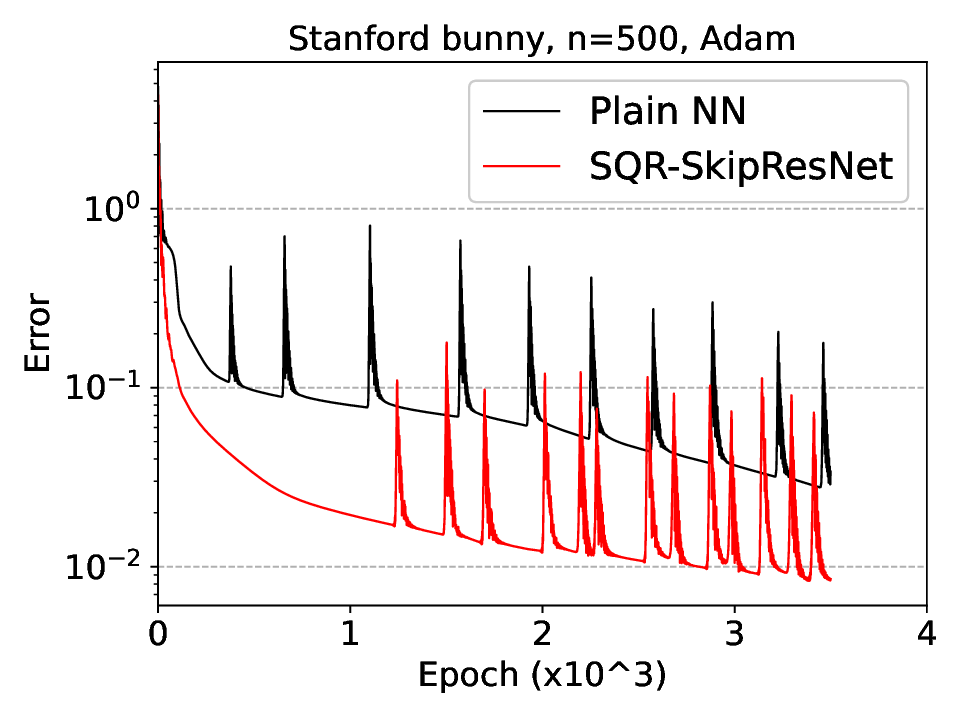}}
\subfigure[]{ \label{Ex7_F4_n500_50x20_pnn_adam}
\includegraphics[width=1.95in]{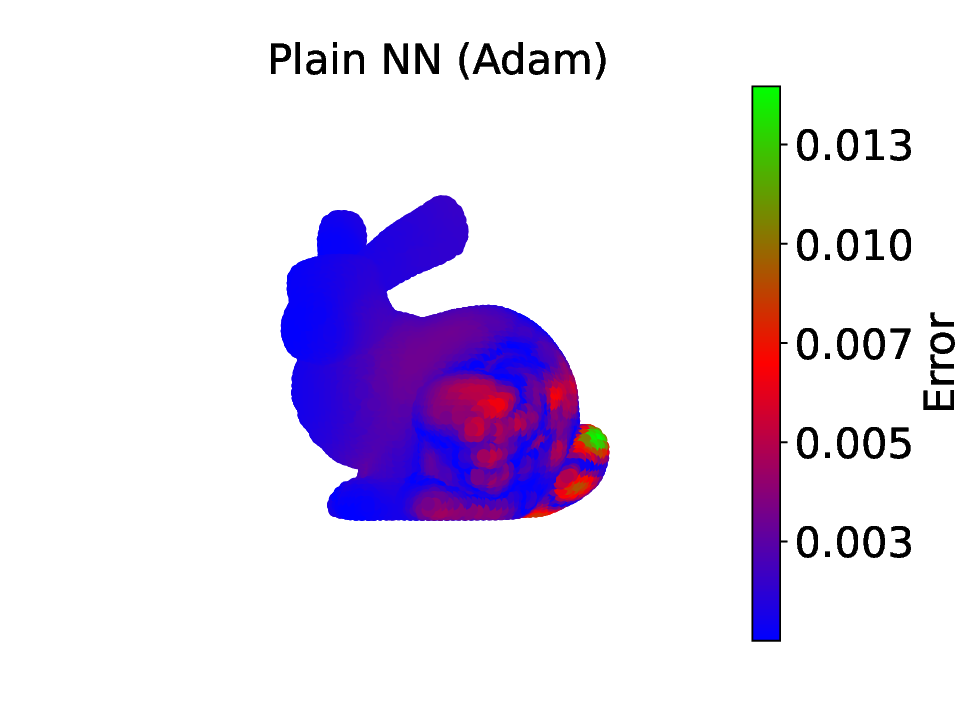}}
\subfigure[]{ \label{Ex7_F4_n500_50x20_sqr_adam}
\includegraphics[width=1.95in]{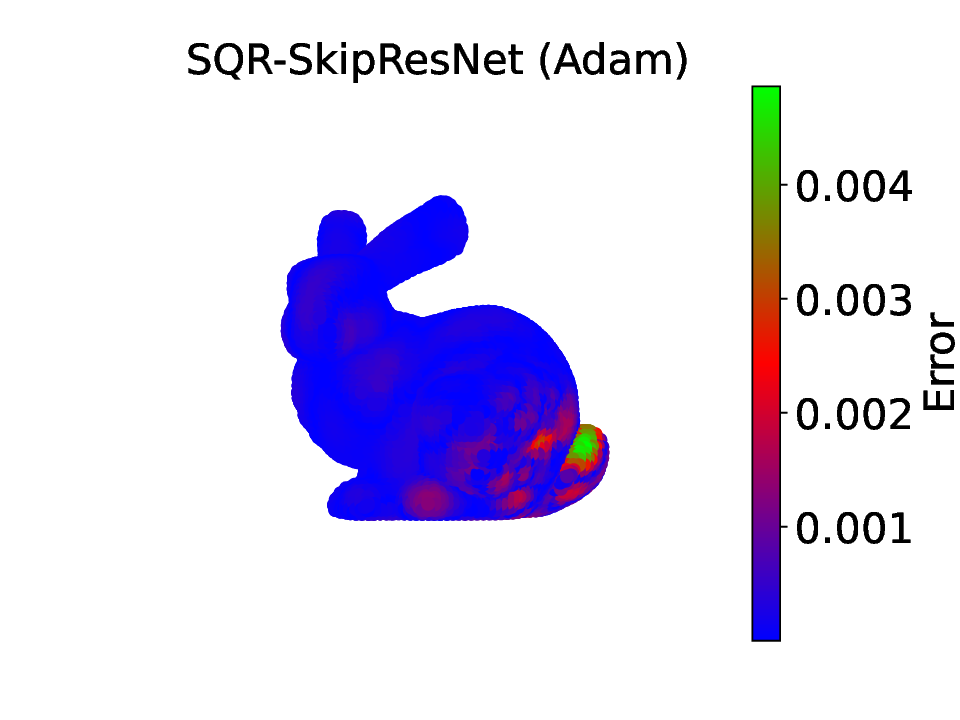}}
\caption{Example 3: Error profile comparison for the Stanford Bunny model using L-BFGS-B (top panel) and Adam optimizers (bottom panel). Training errors are indicated by the dotted line, and validation errors are represented by the solid line.} \label{Ex7_2}
\end{figure}

In Fig.~\ref{Ex7_2}, the training process (dotted line) is depicted with 500 data values, while the remaining 7671 points are reserved for validation error assessment (solid line). The top panel showcases results obtained using the L-BFGS-B optimizer, while the bottom panel displays outcomes achieved through the Adam optimizer. As demonstrated in Fig.~\ref{Ex7_F4_n500_50x20}, the SQR-SkipResNet surpasses the Plain NN in terms of accuracy and convergence rate across both the training and test datasets. The recorded CPU times amount to 35 seconds for Plain NN and 15 seconds for SQR-SkipResNet. Plots (b) and (c) offer insight into the maximum absolute error, highlighting an accuracy improvement of approximately 70\% when implementing the proposed network architecture.

Moreover, the lower panel of the figure reveals that the efficacy of the SQR-SkipResNet method persists even when utilizing the Adam optimizer. Plot (a) illustrates a more rapid convergence rate for the proposed method when evaluated against test data. The plots (b) and (c) portraying the maximum absolute error clearly exhibit significantly improved accuracy achieved through the proposed approach. This consistent superiority serves to highlight the distinct advantages of the SQR-SkipResNet approach over its alternatives. In comparing the L-BFGS-B and Adam optimizers, it becomes evident that the former displays enhanced performance in both accuracy and CPU time, accomplishing the desired accuracy level more efficiently.

\textit{One might wonder about the advantages of employing deep neural networks and their computational implications.} To illustrate this aspect, we emphasize the significance of network depth in neural networks, as shown in Fig.~\ref{Ex7_3}, specifically focusing on F4 with $n=500$ data points and employing the L-BFGS-B optimizer. The results presented here encompass scenarios with 5, 10, and 20 hidden layers, each consisting of 50 neurons.

Examining plot (a), which illustrates the validation error using the Plain NN, we note that increasing the number of hidden layers from 5 to 10 results in a decreased convergence rate. Interestingly, increasing the number of layers to 20, denoted by $n_l=20$, leads to the most favorable convergence rate when compared to the cases of $n_l=5$ and $n_l=10$. Regarding accuracy, variations in the number of layers yield only marginal changes in accuracy. However, the network with 20 hidden layers displays the highest error.

Conversely, in the case of SQR-SkipResNet, a deeper network correlates with improved convergence rate and enhanced accuracy. This suggests that deeper hidden layers can  identify features when embedded within an appropriate neural network architecture for this particular example. This stands in contrast to our findings in the second example (Table~\ref{tab:ex3a}), which highlighted the problem-dependent nature of selecting an optimal number of layers. In this context, the recorded CPU times for models with 5 and 20 hidden layers amount to 19 and 16 seconds, respectively. This observation suggests that deeper networks may not necessarily result in longer CPU times; rather, they can potentially expedite training due to improved convergence rates, as evident in this case.

 \begin{figure}
\centering%
\subfigure[]{ \label{Ex7_F4_n500_50x20x10x5_pnn}
\includegraphics[width=2.95in]{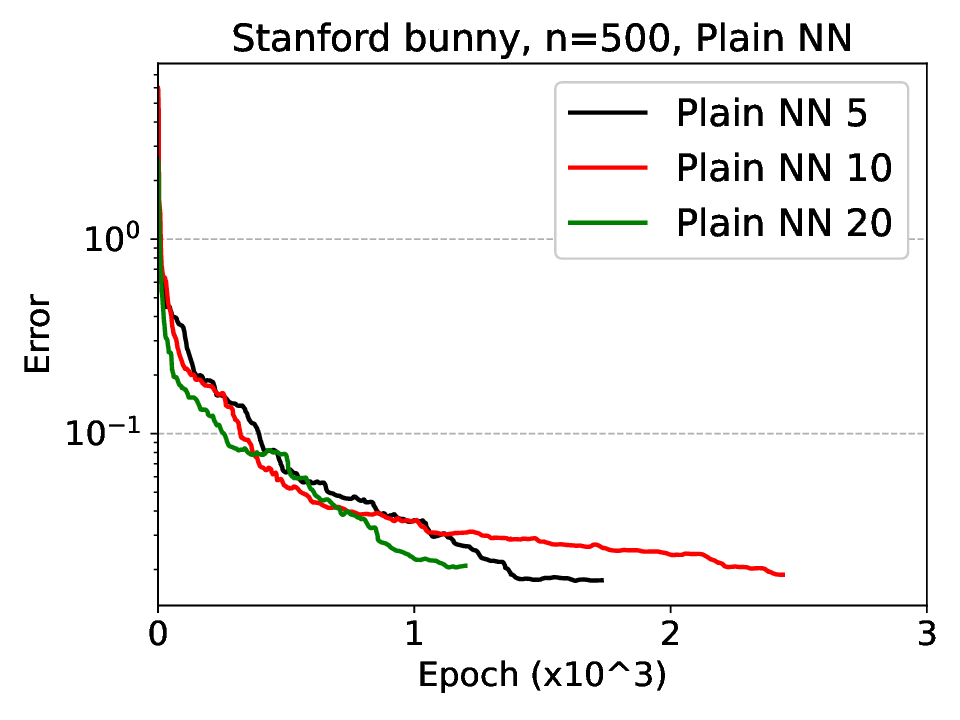}}
\subfigure[]{ \label{Ex7_F4_n500_50x20x10x5_sqr}
\includegraphics[width=2.95in]{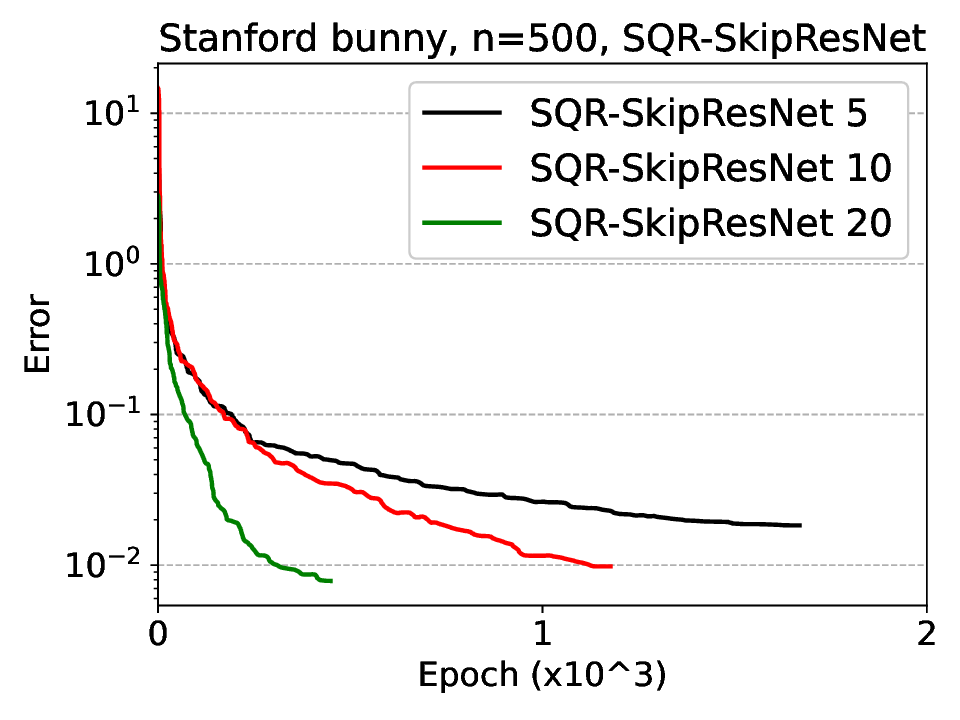}}
\caption{Example 3: Profiles of the validation errors for interpolating the the Stanford Bunny for different number of layers using (a) Plain NN and (b) SQR-SkipResNet.} \label{Ex7_3}
\end{figure}

}
\end{example}

%%%%%%%%%%%%%%%%%%%%%%%%%%%%%%%%%%%%%%%%%%%%%%%%%%%
%%%%%%%%%%%%%%%%%%%%%%%%%%%%%%%%%%%%%%%%%%%%%%%%%
%%%%%%%%%%%%%%%%%%%%%%%%%%%%%%%%%%%%%%%%%%%%%%%%%%%%
%%%%Example 5
\begin{example}\rm{
In our final example, we delve into the performance evaluation of the proposed SQR-SkipResNet for solving the inverse problem, specifically focusing on the Burgers' equation. The ground truth coefficients are $\lambda_1 = 1$ and $\lambda_2 = \nu = \frac{1}{100\pi} = 0.003183$, while the initial estimates are $\lambda_1=2.0$ and $\lambda_2=0.2$. The outcomes of the investigation are presented in Figure~\ref{Ex8_1}, which showcases the results obtained during training and validation for $n=500$ using the L-BFGS-B optimizer.

Further analysis is conducted for different network architectures. Figure~\ref{Ex8_n500_50x10} demonstrates the outcomes for the configuration $(n_l,n_n)=(10,50)$, revealing improved accuracy for both collocation and validation data when employing the proposed method. The predicted values of $\lambda_1$ using SQR-SkipResNet and Plain NN show errors of 0.25\% and 0.35\%, respectively, when compared with the exact results. Additionally, the percentage errors for predicting $\lambda_1$ and $\lambda_2$ are 1.55\% and 3.29\% for SQR-SkipResNet and Plain NN, respectively.
 Extending this analysis, Fig.~\ref{Ex8_n500_50x20} showcases the results for $(n_l,n_n)=(20,50)$. It is evident that a deeper network architecture leads to enhanced accuracy when utilizing the proposed method. Notably, as the number of hidden layers increases, Plain NN demonstrates larger errors.
This effect is more pronounced in Fig.~\ref{Ex8_n500_50x50}, which presents the results for a large number of hidden layers $(n_l=50)$. Consequently, we can conclude that the proposed neural network architecture not only improves accuracy but also exhibits greater stability concerning varying numbers of hidden layers.

Comparing the two plots, we observe that the accuracy difference between Plain NN and SQR-SkipResNet becomes more pronounced as the network size increases. This emphasizes the crucial role of architecture selection in achieving stable results.

\begin{figure}
\centering%
\subfigure[$(n_l,n_n)=(10,50)$.]{ \label{Ex8_n500_50x10}
\includegraphics[width=1.95in]{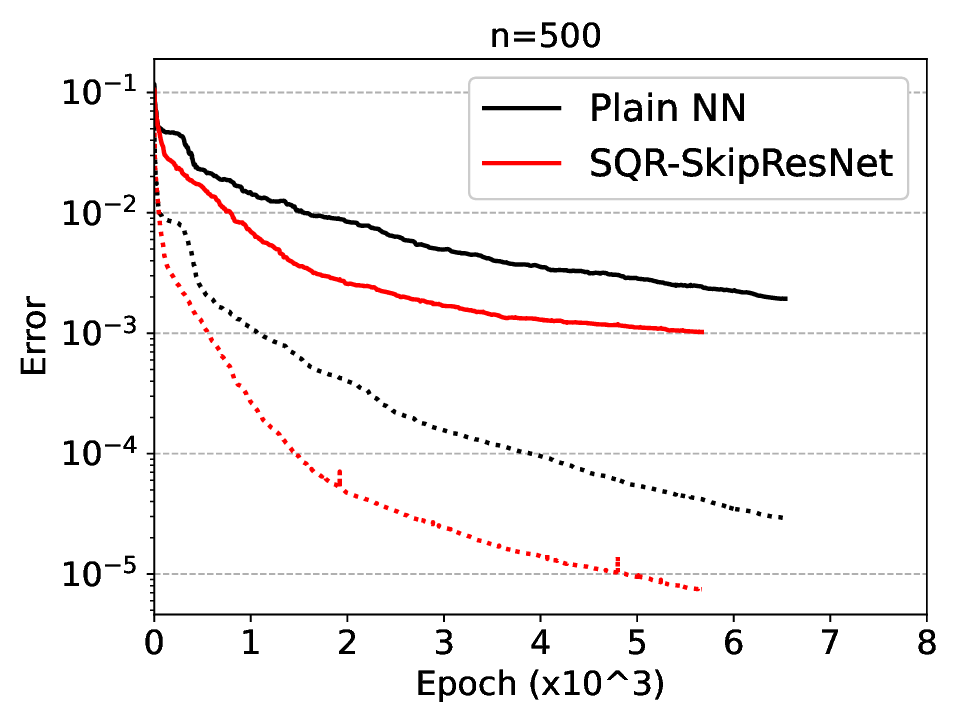}}
\subfigure[$(n_l,n_n)=(20,50)$.]{ \label{Ex8_n500_50x20}
\includegraphics[width=1.95in]{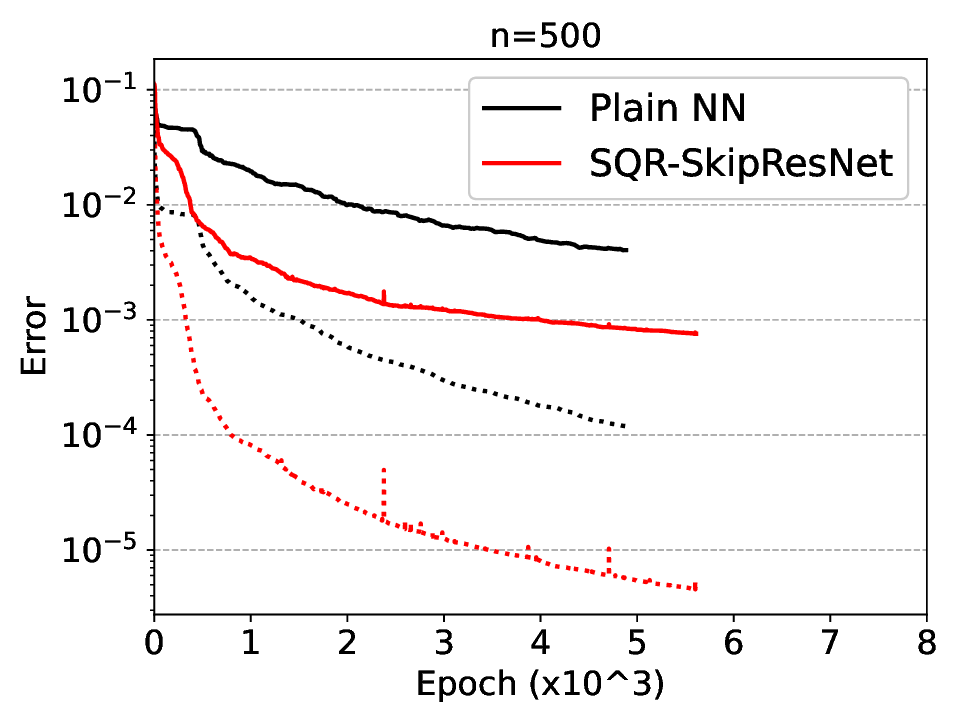}}
\subfigure[$(n_l,n_n)=(50,50)$.]{ \label{Ex8_n500_50x50}
\includegraphics[width=1.95in]{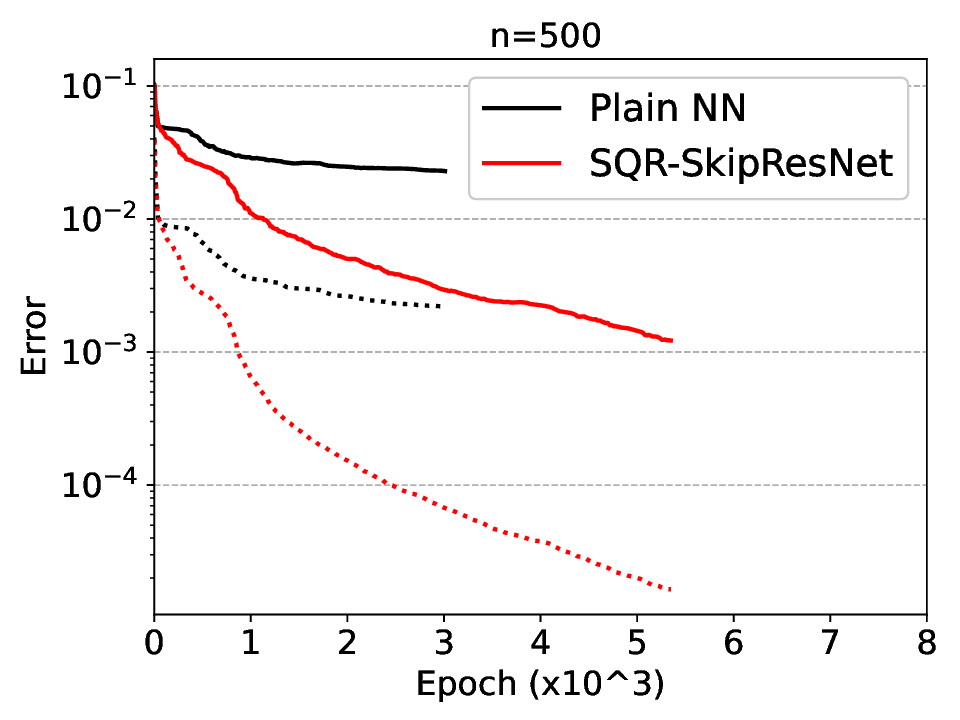}}
\caption{Example 4: Profiles of training (dotted line) and validation error (solid line) for different number of layers.} \label{Ex8_1}
\end{figure}

}
\end{example}

\section{Conclusion}

Throughout this study, we conducted a series of experiments to assess how different neural network setups, including Plain NN and SQR-SkipResNet, perform when it comes to interpolating both smooth and complex functions.
The Plain NN's failure is attributed to significant fluctuations or increasing rate in weight norms, destabilizing the learning process and hindering convergence. Such erratic behavior prevents the network from reaching a satisfactory solution, underscoring the importance of stable weight updates in training neural networks. In contrast, the proposed SQR-SkipResNet exhibits better performance by maintaining stable weight updates and achieving faster convergence.
 Our findings consistently showed that SQR-SkipResNet outperforms other architectures in terms of accuracy. This was especially evident when dealing with non-smooth functions, where SQR-SkipResNet displayed improved accuracy, although it might take slightly more time to converge.
We also applied our approach to real-world examples, like interpolating the shape of a volcano and the Stanford bunny. In both cases, SQR-SkipResNet exhibited better accuracy, convergence, and computational time compared to Plain NN.

Additionally, choosing a deeper network can sometimes decrease accuracy for both Plain NN and SQR-SkipResNet, but we found this depends on the specific problem. For instance, when dealing with the complicated geometry of the Stanford Bunny and its smooth function, we noticed that deeper networks yielded enhanced accuracy, quicker convergence, and improved CPU efficiency. Regardless of whether deeper networks are suitable, the proposed method demonstrated superior performance. As the effectiveness of network depth varies based on the problem, our approach offers a more favorable architecture choice for networks of different depths.

In physics-informed neural networks, where a physics constraint follows the neural network, enhancing the neural network improves overall performance. The proposed method boosts function approximation accuracy in neural networks, suggesting it will enhance total performance in PINN architecture. Testing on an inverse problem using physics-informed neural networks revealed significant accuracy and stability gains with SQR-SkipResNet across various hidden layer configurations, unlike Plain NN. Future research can explore applying this method to tackle more complex PDEs, both forward and inverse, in addition to providing mathematical expressions to support our proposed method.

\section*{Acknowledgments}
Authors gratefully acknowledge the financial support of the Ministry of Science and Technology (MOST) of Taiwan under grant numbers 
112-2221-E-002-097-MY3 and 112-2811-E-002-020-MY3.
We also want to acknowledge the resources and support from the National Center for Research on Earthquake Engineering (NCREE), the NTUCE-NCREE Joint Artificial Intelligence Research Center, and the National Center of High-performance Computing (NCHC) in Taiwan.

\end{document}